\documentclass{svmult}

\usepackage{makeidx}         
\usepackage{graphicx}        
\usepackage{multicol}        
\usepackage[bottom]{footmisc}

\usepackage{amssymb}
\usepackage{subfigure}

\makeindex             

\begin{document}

\title*{Face Recognition Based on Sequence of Images}
\author{Jacek Komorowski\inst{1}\and
Przemyslaw Rokita\inst{2}}
\institute{
Military University of Technology,
Faculty of Cybernetics,
Warsaw, Poland
\texttt{jac99@o2.pl}
\and 
Military University of Technology,
Faculty of Cybernetics,
Warsaw, Poland
\texttt{p.rokita@ii.pw.edu.pl}
}
%
%
\maketitle

\index{Komorowski J.(1)} 
\index{Rokita P.(2)} 

\abstract{
This paper presents a face recognition method based on a sequence of images.
Face shape is reconstructed from images using a combination of structure-from-motion and multi-view
stereo methods.
The reconstructed 3D face model is compared against models held in a gallery.
The novel element in the presented approach is the fact, that the reconstruction is based only on input images and doesn't require a generic, deformable face model.
Experimental verification of the proposed method is also included.
}

\section{Introduction}

Three dimensional face recognition is an active and growing field of research \cite{89:bow} \cite{89:chang}.
Using spatial information allows to mitigate some of the problems faced by methods based solely on visual information. 3D face recognition methods are less dependent on face pose and lighting variations.
One of the barriers to a mass deployment of this technology is a difficulty with a face shape acqusition.
Active vision techniques, such as laser scanning, are not appropriate for practical usage. Laser scanners are rather large, expensive and may be damaging to human eyes.
Alternative, passive techniques, such as stereovision, multi-view stereo or structure-form-motion, are not very well suited for human face shape reconstruction. These methods are based on finding corresponding points on multiple images, that is points which are projections of the same scene point. Human skin has a relatively homogeneous texture which makes an automatic matching a difficult task.

Majority of methods which use passive vision techniques for face shape reconstruction, either uses complex image acquision setup (e.g. set of 5 cameras \cite{89:spre}) or utilises a generic, deformable face models (e.g. \cite{89:cheng}).
Complex camera setups complicate practical deployment.
Model-based approach is criticized \cite{89:fid}, that it doesn't allow to model subtle details important for accurate face recognition, as reconstruction result are limited by a model parameter space.

The method presented in this paper uses a sequence of images from a single camera. Therefore it's easy to use as there's no need for a complicated equipment. Additionally it's based solely on input images and doesn't require a generic face model.
Multi-view stereo algorithms can be used to reconstruct a 3D object model from a set or sequence of images taken from multiple viewpoint.
Over the last years a significant progress was made in this area and a number of high-quality algorithms were developed.
Best methods reviewed in 
\cite{89:seitz}
can deal with very demanding scenarios, where input images depict objects with little texture, containing few points which can be automatically matched across multiple images.
For very demanding DinoRing 
\footnote{\url{http://vision.middlebury.edu/mview/data}}
test set, containing images of a plaster dinosaur taken from multiple viewpoints, the best algorithms surveyed in \cite{89:seitz} were able to reconstruct over 90\% of the object surface with error below $0.4$ mm.
Unfortunately multi-view stereo algorithms assume that all images are fully calibrated, that is both intrinsic (camera focal length, distortion coefficients) and extrinsic (camera pose) parameters for each image are known.
Such algorithms cannot be used  when a sequence contains images of an object moving freely in front of the camera.
Intrinsic camera parameters are fixed, and can be estimated with a prior calibration.
But extrinsic parameters are different for each image and cannot be easily estimated.
To use some high-quality multiview-stereo algorithm for face shape reconstruction from a sequence of images, extrinsic parameters for each image in the sequence must be estimated.

\section{Details of the method}

\begin{figure}
\centering
\subfigure{
\includegraphics[height=2.4 cm]{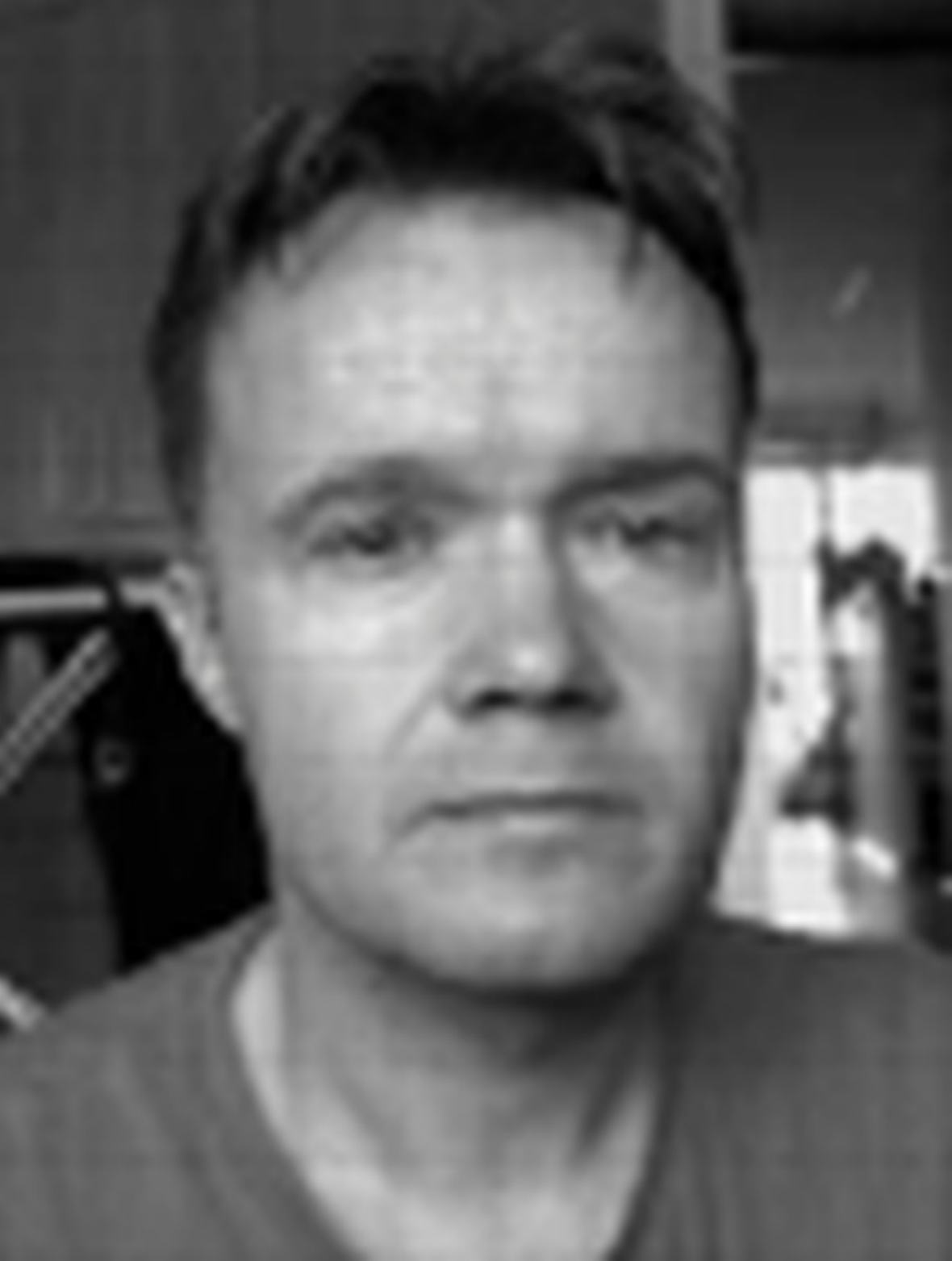}
}
\subfigure{
\includegraphics[height=2.4 cm]{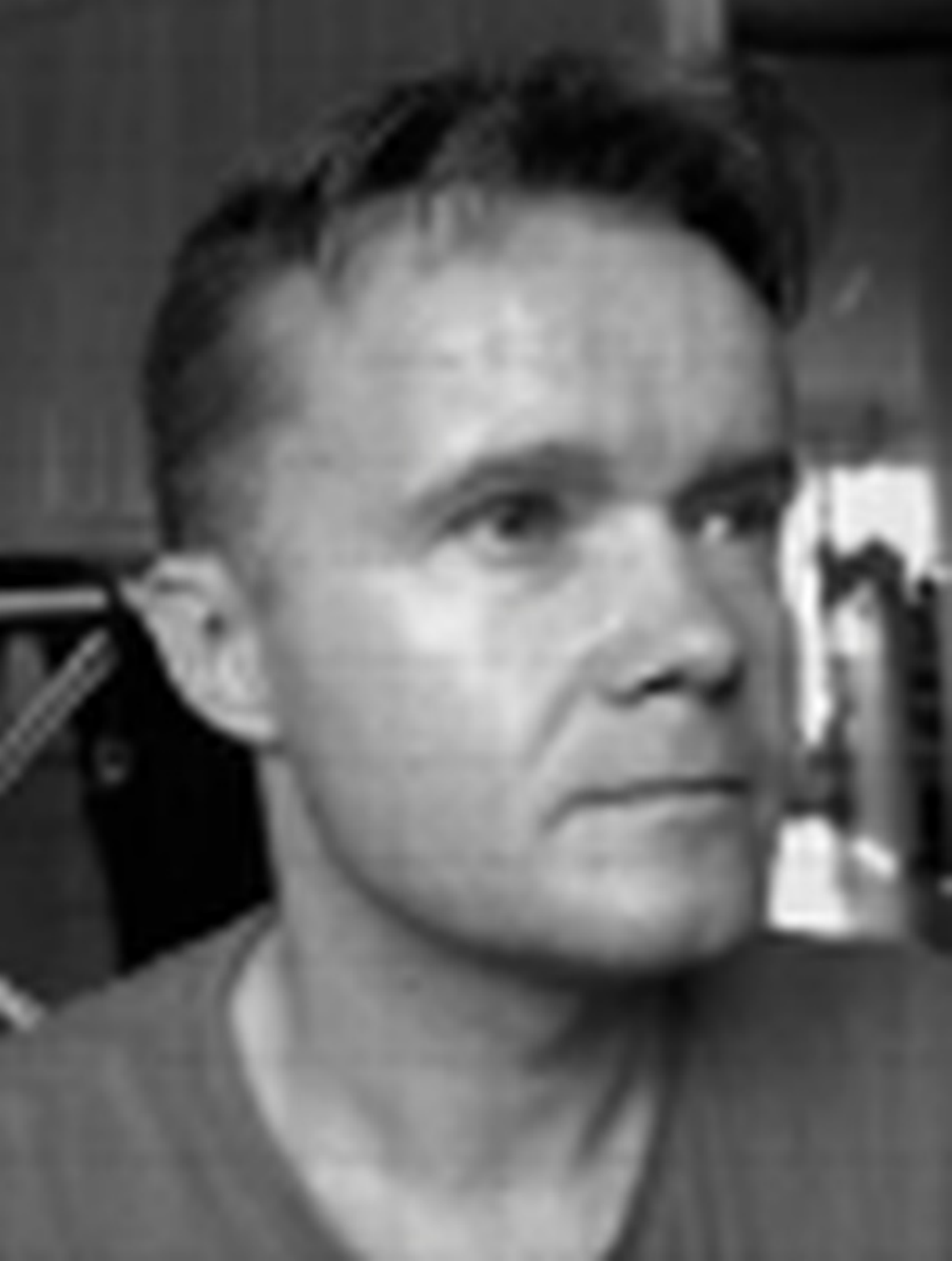}
}
\subfigure{
\includegraphics[height=2.4 cm]{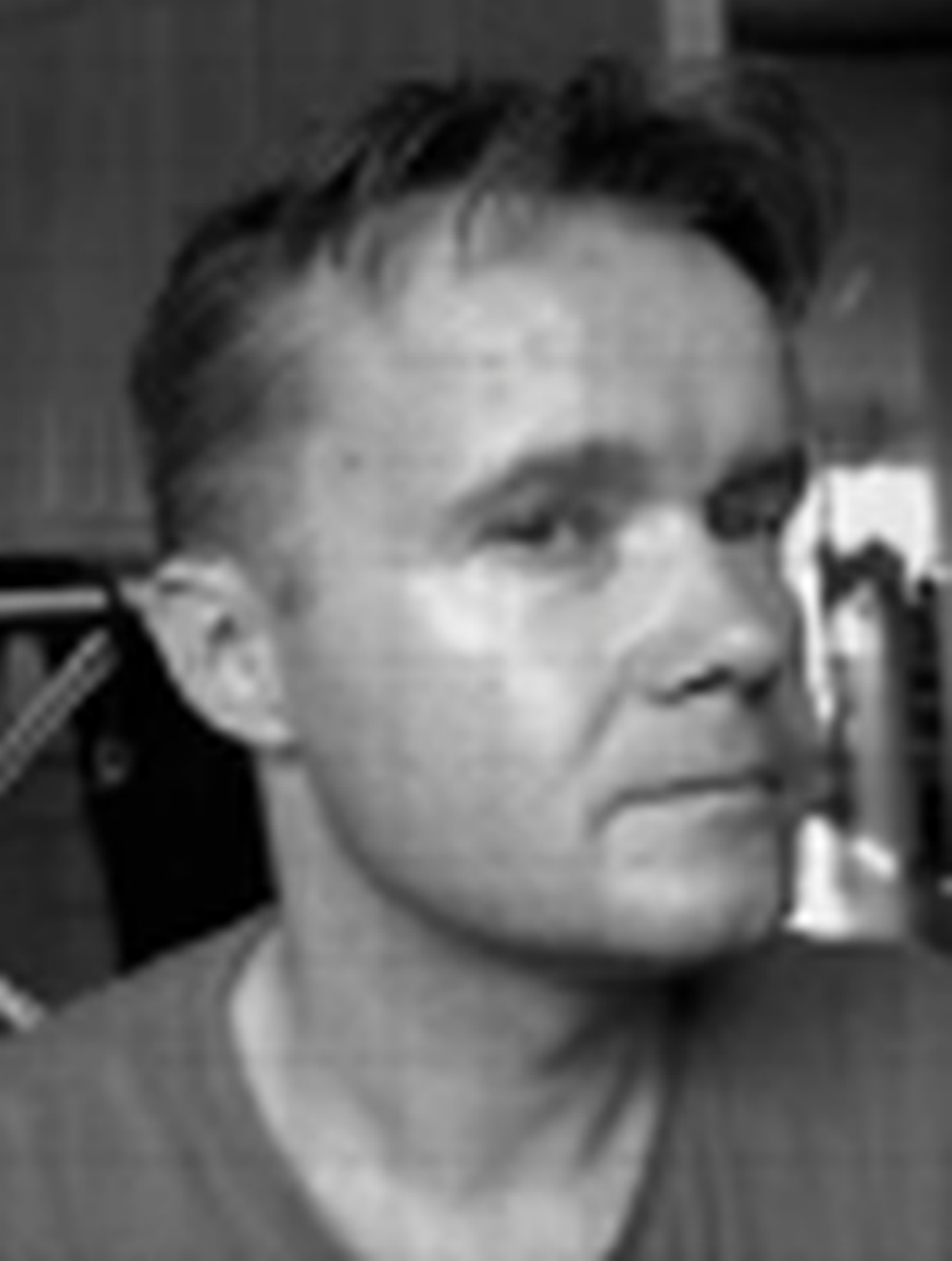}
}
\subfigure{
\includegraphics[height=2.4 cm]{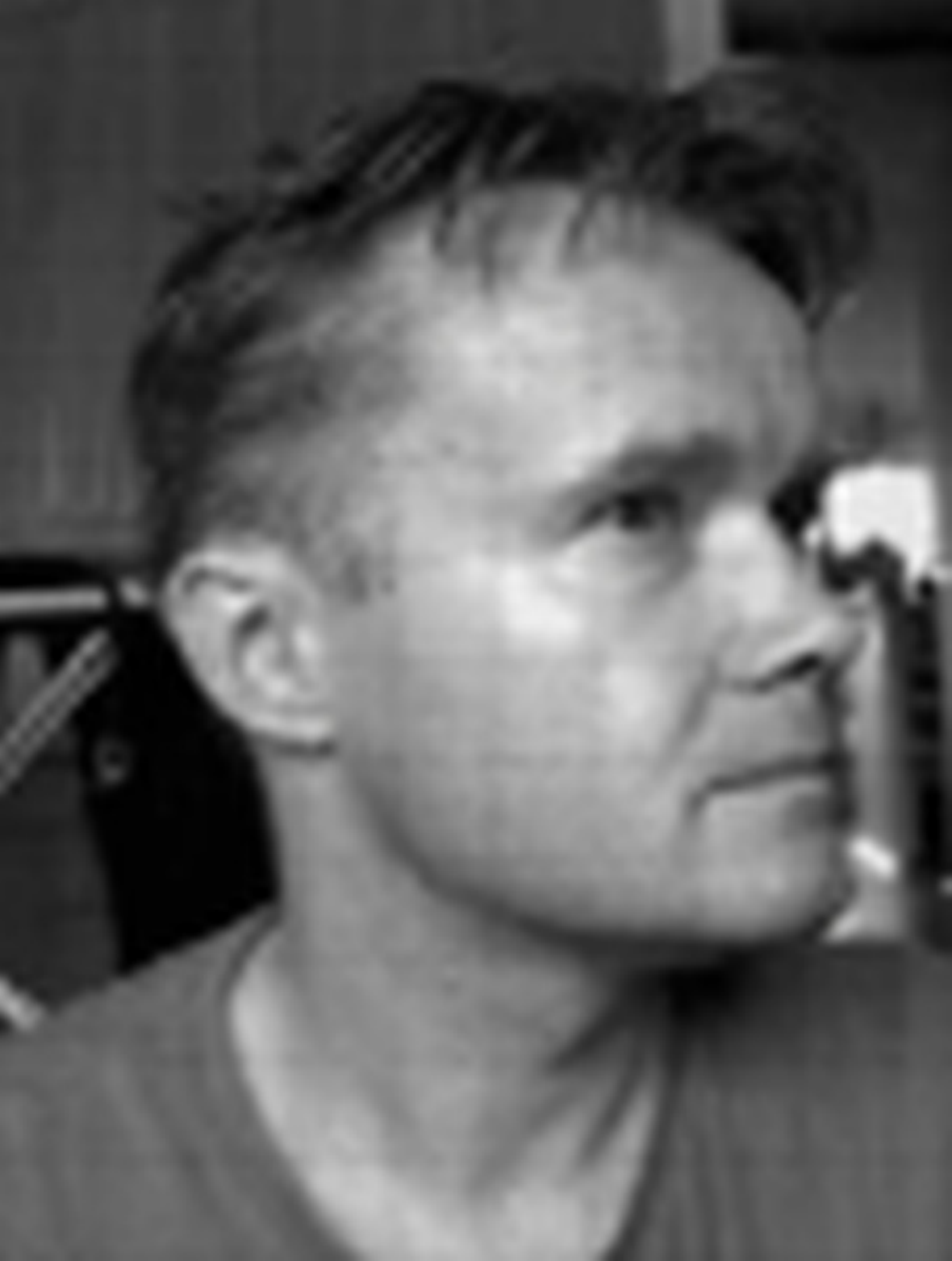}
}
\subfigure{
\includegraphics[height=2.4 cm]{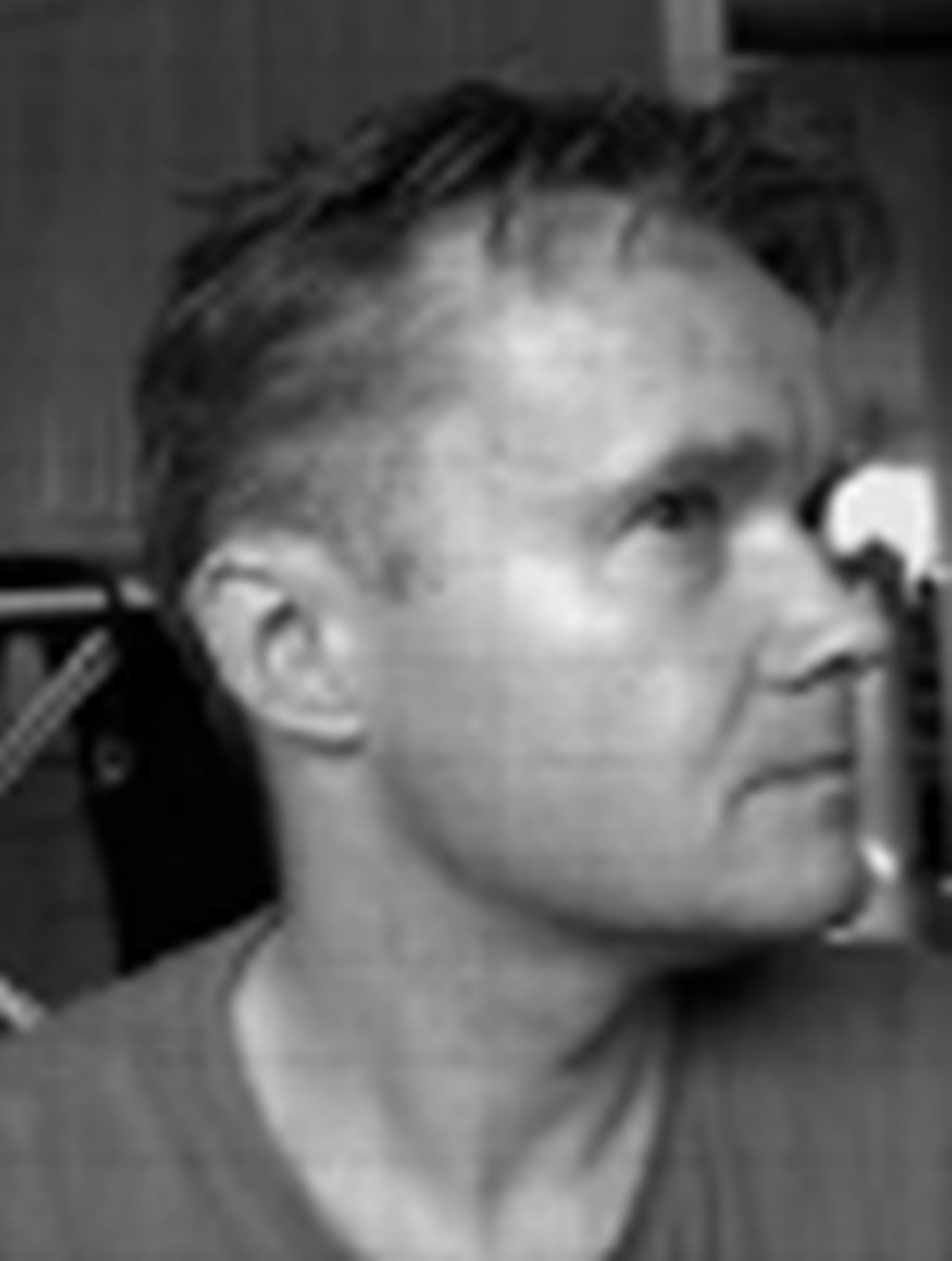}
}
\caption{Exemplary input sequence (5 from 80 images).} 
\label{89:4:dane_wej}
\end{figure}

This section describes details of our face recognition method. The method is based on a sequence of images from a monocular camera.
It's assumed that a person sits in front of the camera and is asked to rotate his head left and right.
An exemplary input sequence is depicted on Fig. \ref{89:4:dane_wej}.

\begin{figure}
\centering
\includegraphics[height=4.5 cm]{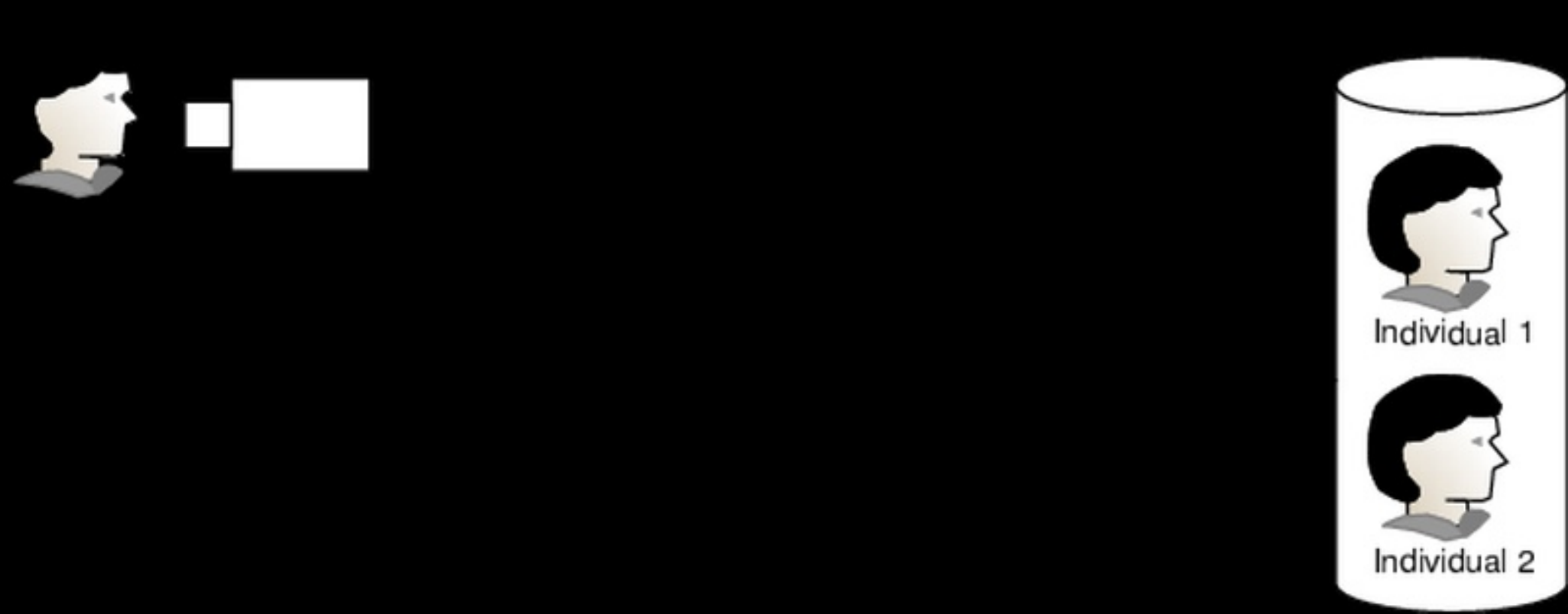}
\caption{Recognition system concept}
\label{89:2}
\end{figure}

Main steps of the presented method are depicted on Fig. \ref{89:2}.

\begin{figure}
\centering
\subfigure{
\includegraphics[height=3.4 cm]{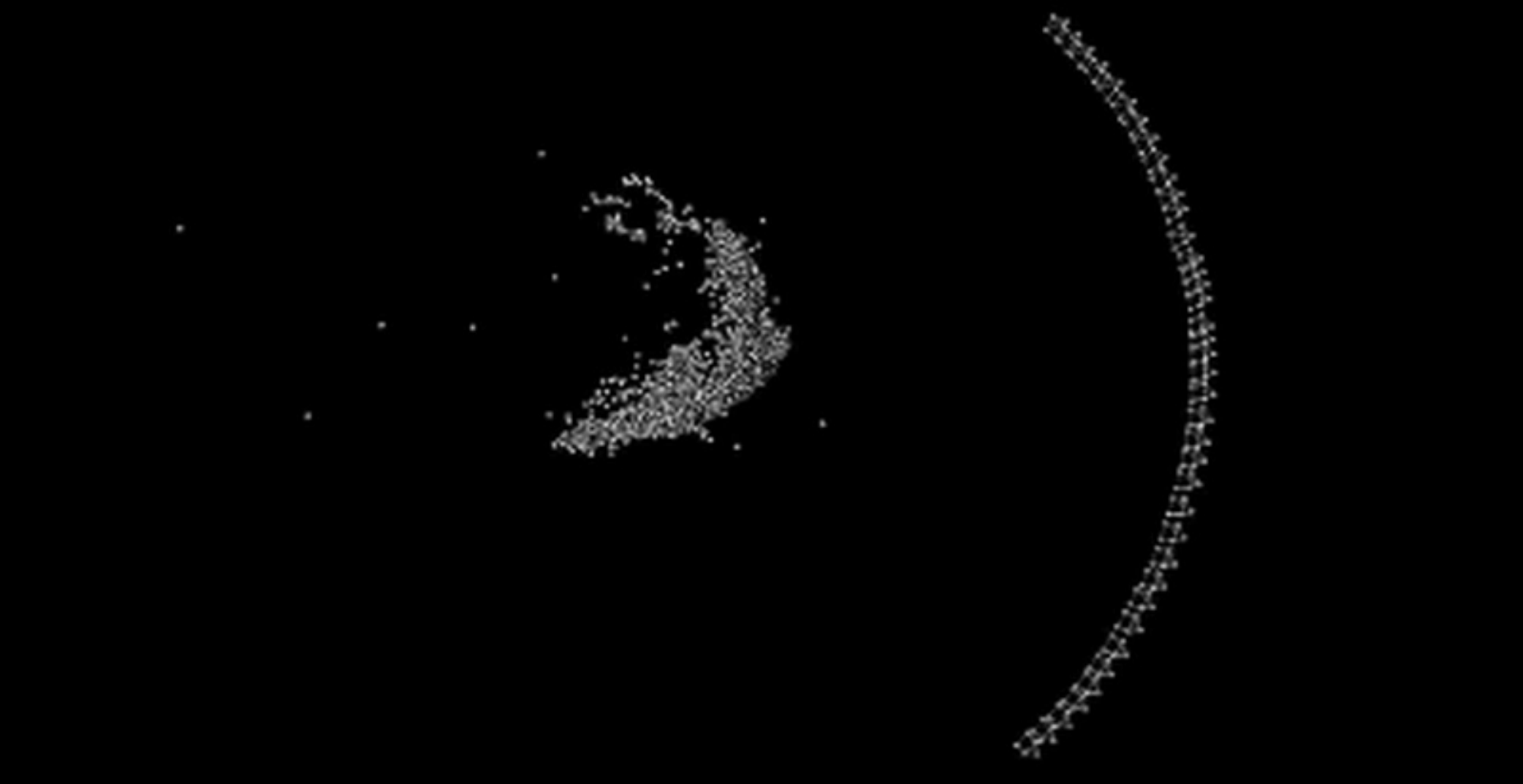}
} \\
\subfigure{
\includegraphics[height=3.4 cm]{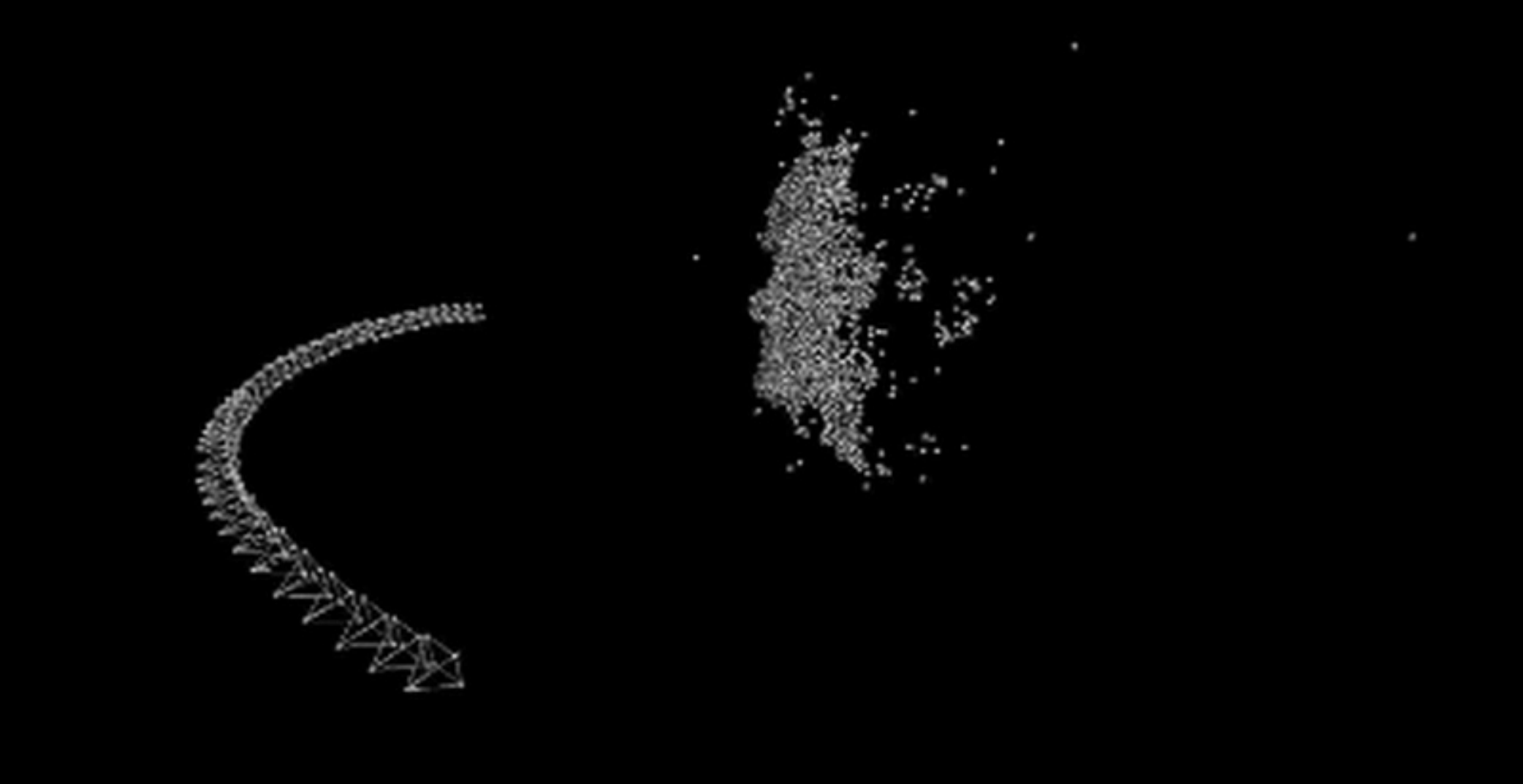}
}
\caption{Estimated camera poses (pyramids) and a sparse face model (point cloud) based on a sequence of images from Fig. \ref{89:4:dane_wej}.} 
\label{89:4:krok4f_1}
\end{figure}

\paragraph{Step 1} Extrinsic camera parameters (rotation matrix $\mathbf{R}$ and translation vector $\mathbf{T}$) are estimated for each image in the sequence. 
This is done using a method developed by authors and described in \cite{89:kom1} and \cite{89:kom2}.
The method is designed to work well with demanding scenarios, where input images contain little texture.
It doesn't use a generic, deformable face model and is based solely on input data.
Results of this step are depicted on Fig. \ref{89:4:krok4f_1}.	

\begin{figure}
\centering
\includegraphics[width=2 cm]{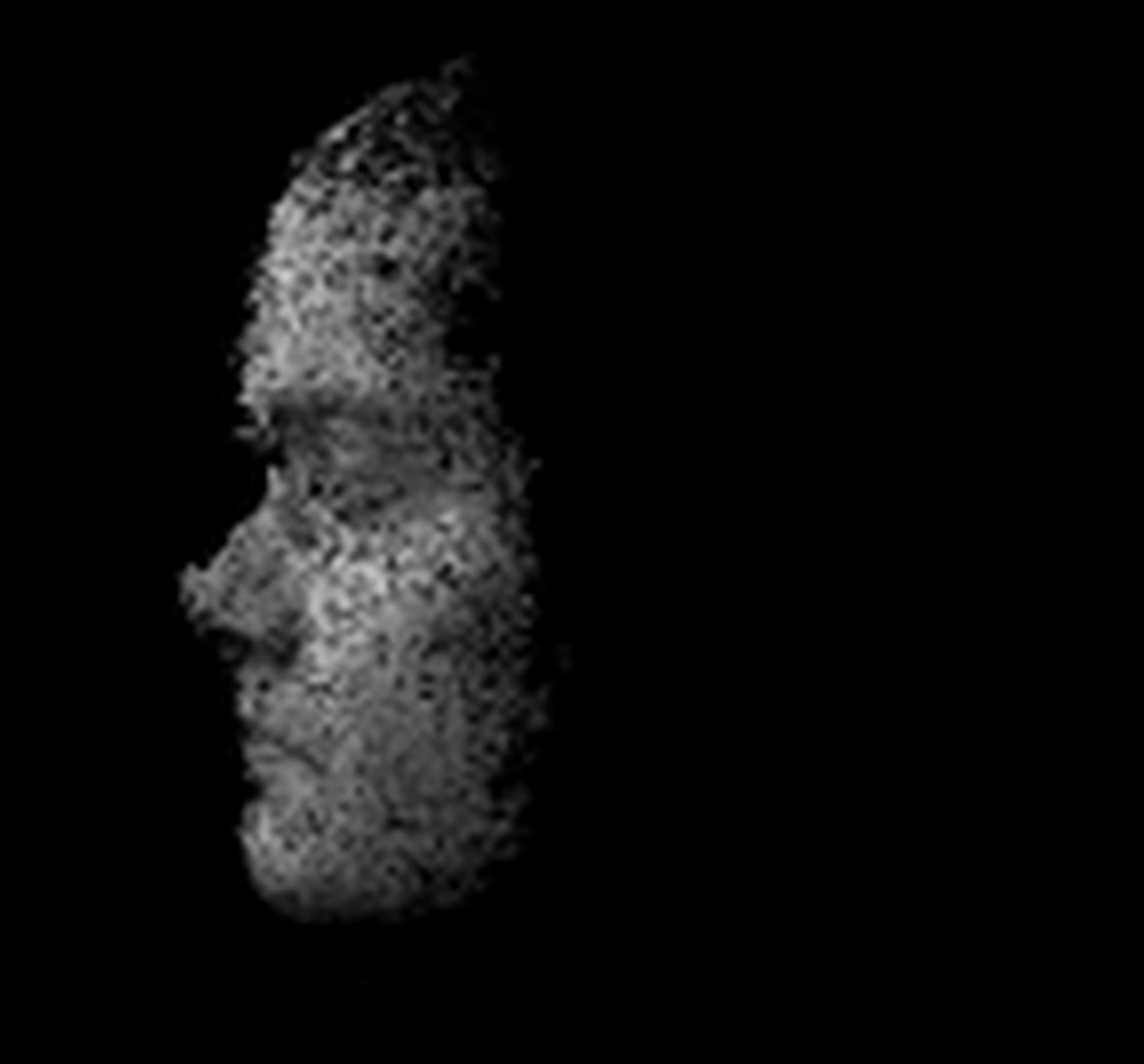}
\includegraphics[width=2 cm]{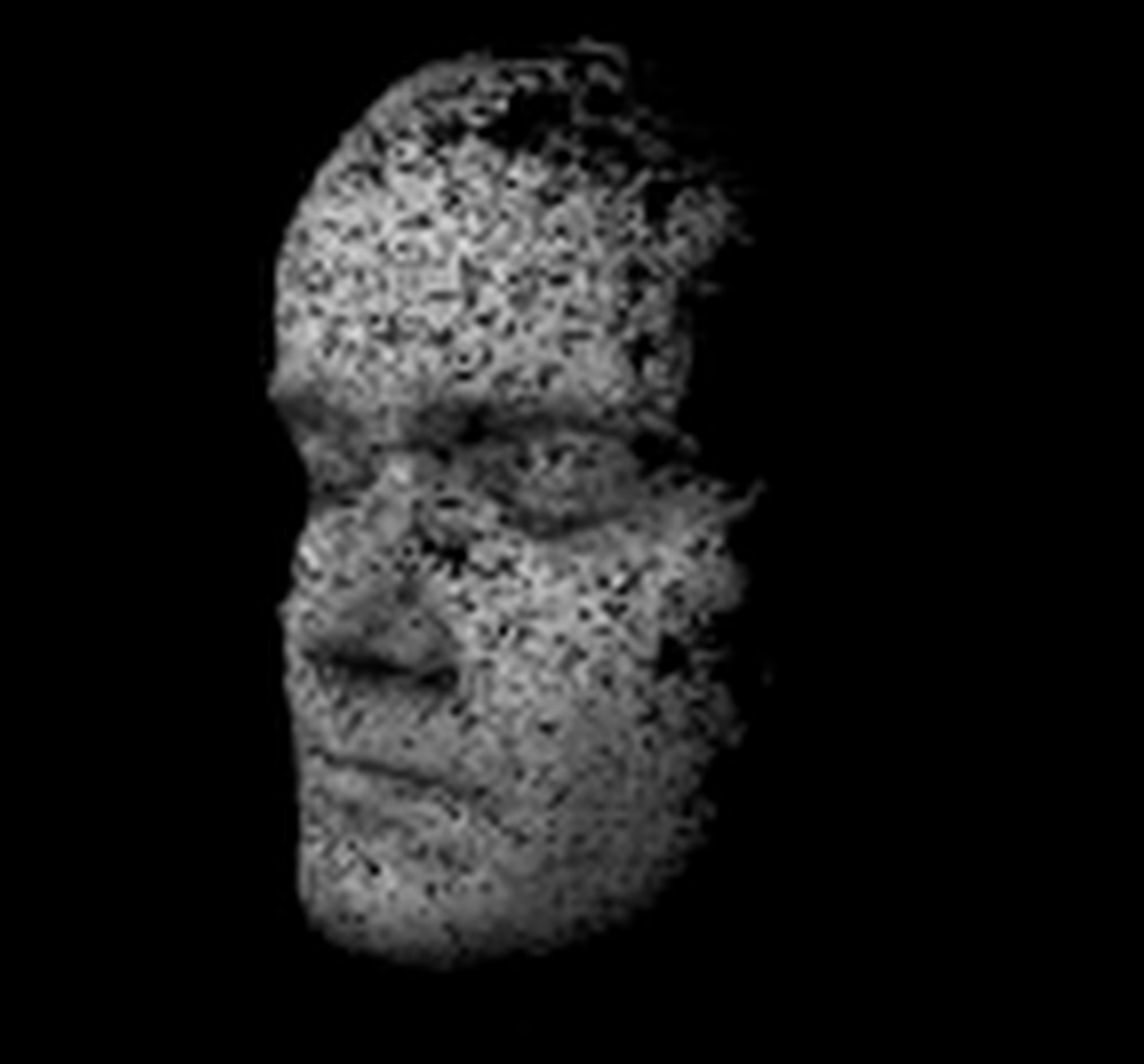}
\includegraphics[width=2 cm]{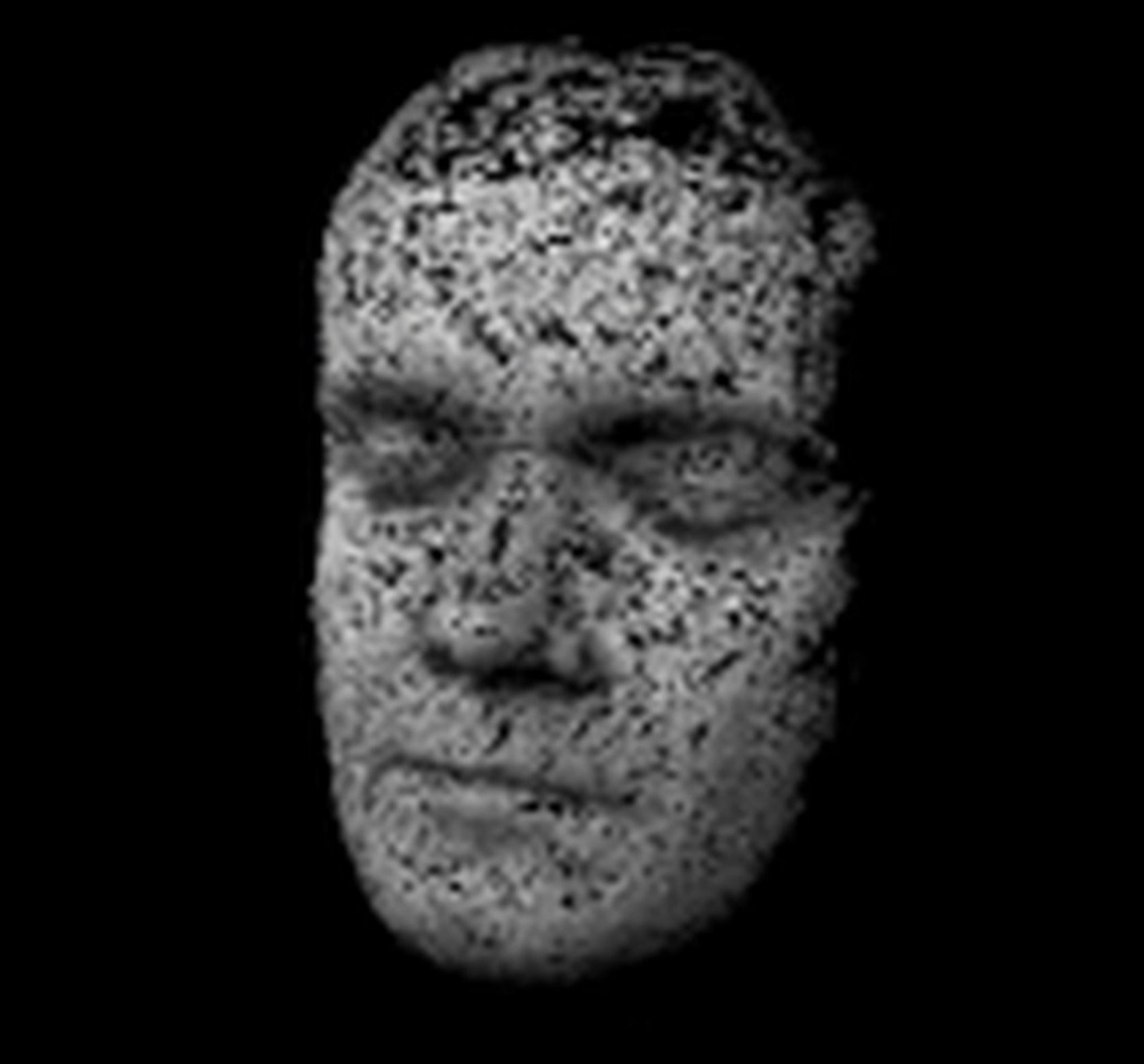}
\includegraphics[width=2 cm]{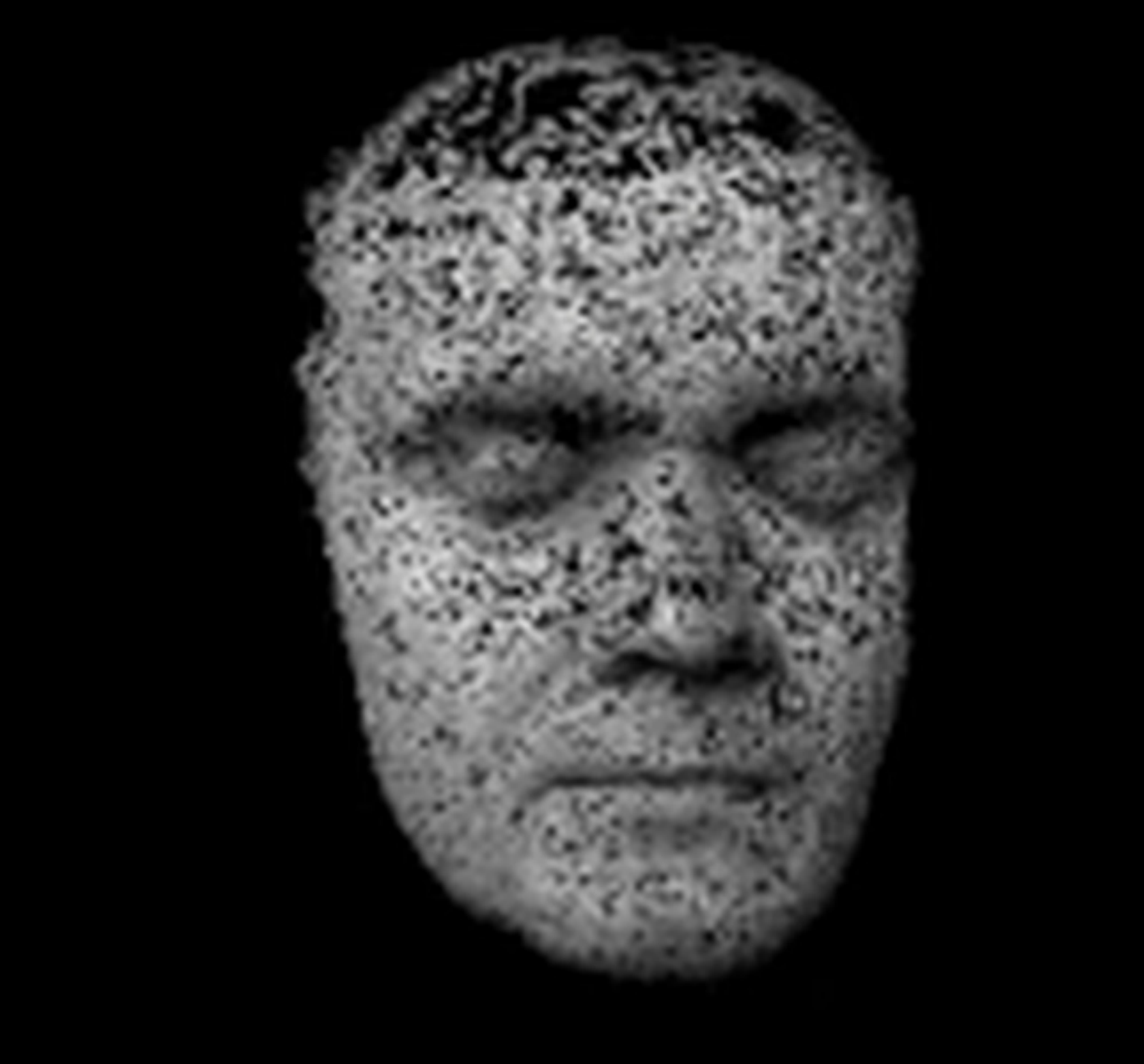}
\includegraphics[width=2 cm]{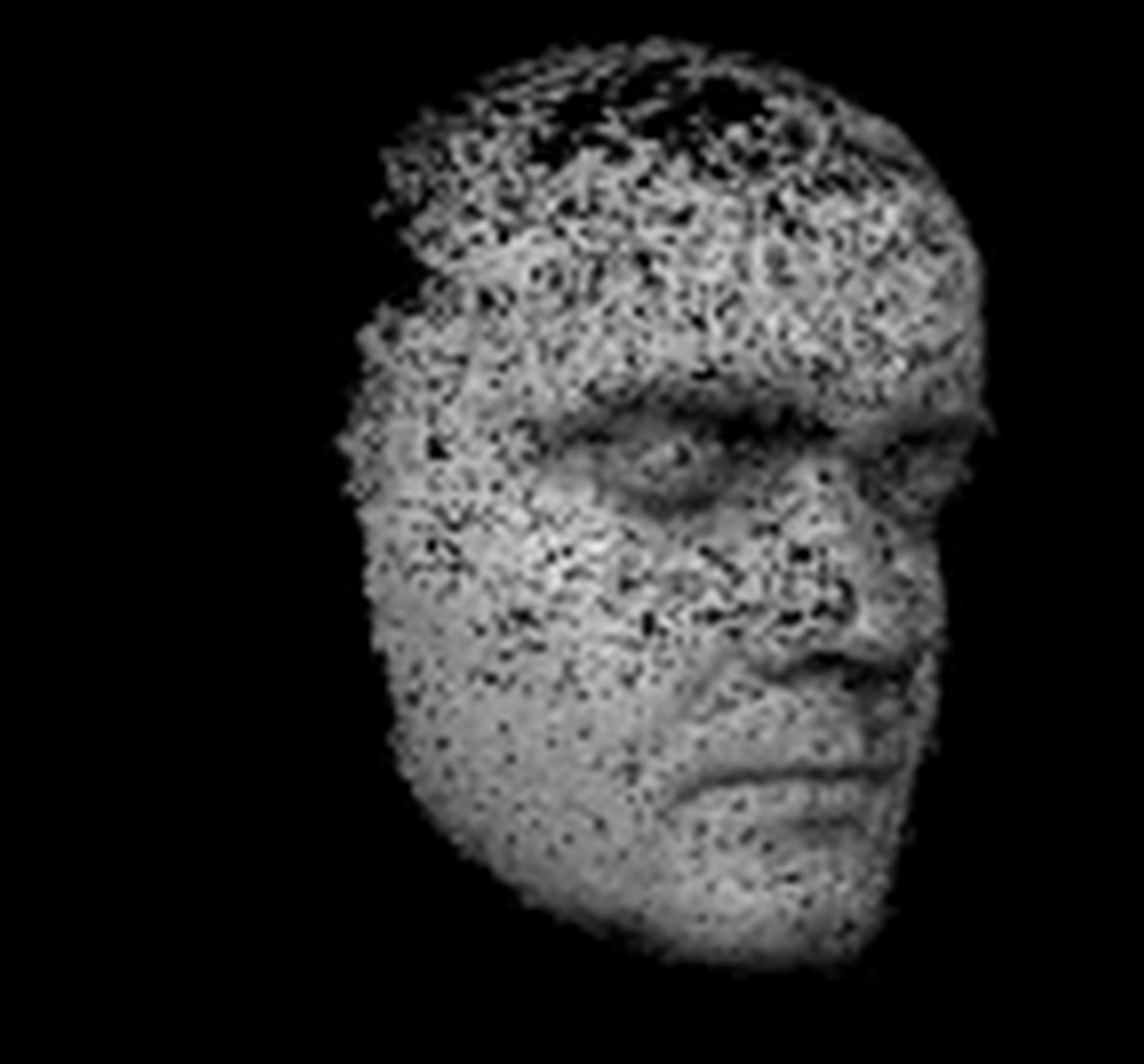}
\includegraphics[width=2 cm]{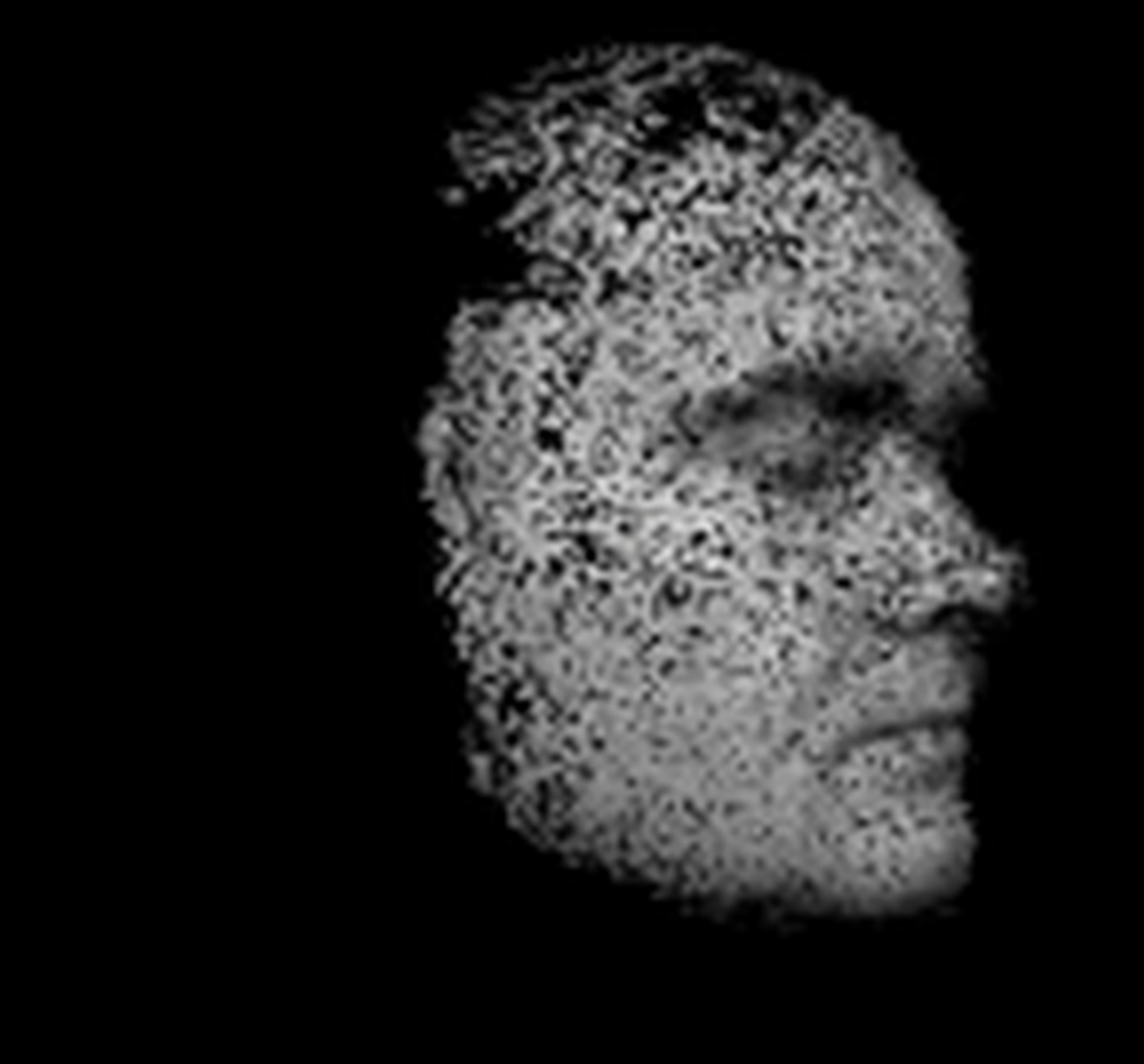}
\includegraphics[width=2 cm]{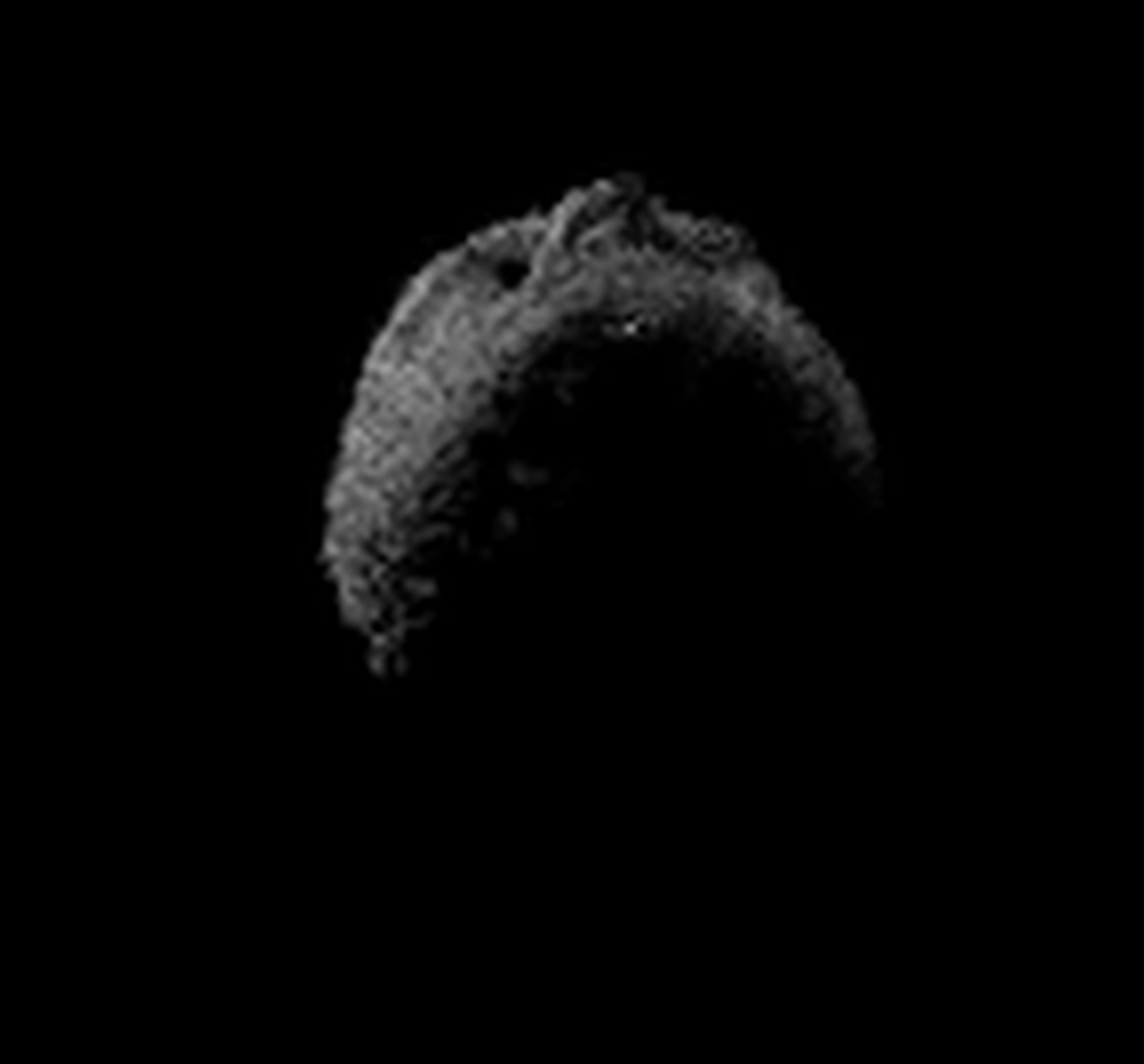}
\includegraphics[width=2 cm]{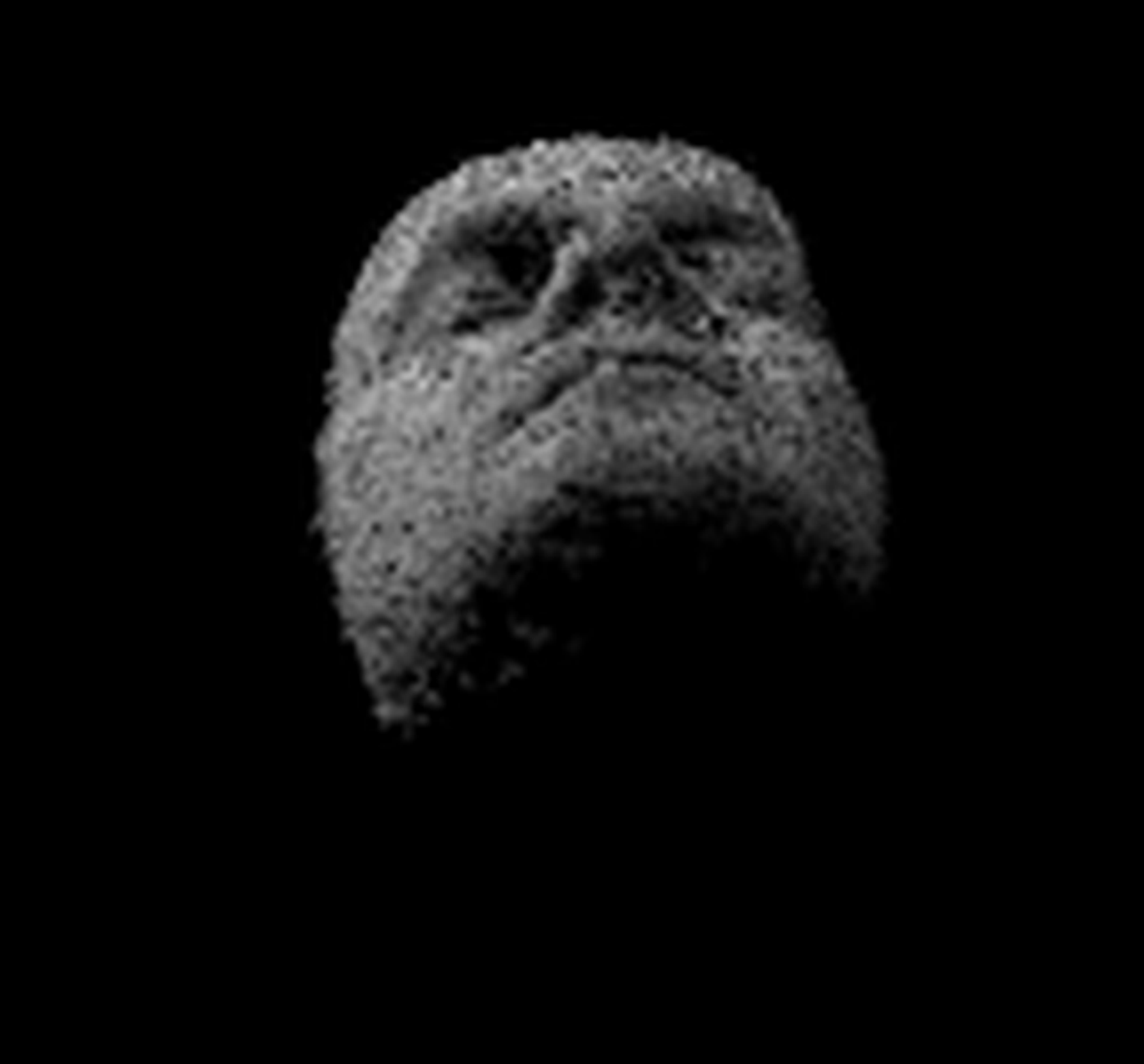}
\includegraphics[width=2 cm]{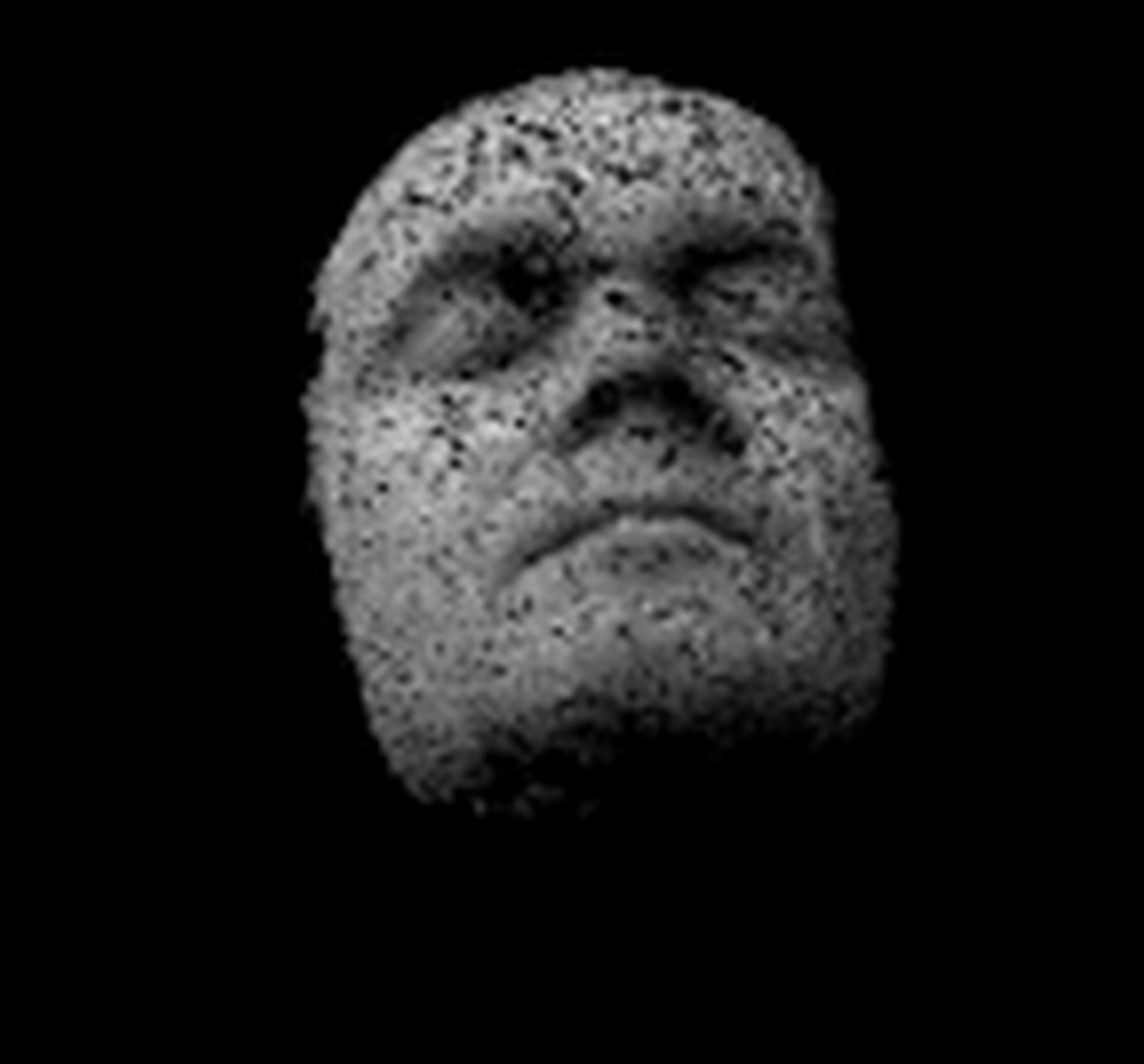}
\caption{Face reconstruction results based on a sequence from Fig. \ref{89:4:dane_wej}} 
\label{89:3:reconst}
\end{figure}

\paragraph{Step 2} 
Once camera extrinsic parameters are estimated, any multi-view stereo algorithm can be used to reconstruct a 3D face shape.
In our implementation a patch-based multi-view stereo method PMVS \cite{89:fur}
\footnote{\url{http://grail.cs.washington.edu/software/pmvs/}} 
was used.
An input to the PMVS algorithm is a sequence of images and estimated camera extrinsic parameters. 
The output is a cloud of oriented points (see Fig. \ref{89:3:reconst}).

\paragraph{Step 3}
Face model reconstructed from an image sequence is compared with models in a gallery.
Distance between point clouds is used as a similarity measure between two face models.
Distance between two point clouds is defined as an average Euclidean distance between each point from the first model
to the closest point in the second model.
Two face models usually do not fully overlap. Due to differences in input sequences
\footnote{E.g. different maximum face rotation angle.}
one cloud may contain regions from a reconstructed object surface, not  presented in the second model.
To deal with this problem a relatively simple heuristic is used.
A median distance between each point from the first model and the closest point in the second model is calculated, and points further away than some small multiple of the median are discarded.
Formal definition of the distance metric used to compare 2 point clouds is as follows:

Let $\mathcal{C}_1 \subset \mathbb{R}^3$ and $\mathcal{C}_2 \subset \mathbb{R}^3$ denote two clouds consisting of points in 3D Cartesian space. 
$d \left( p, \mathcal{C} \right)$ denotes a distance of a point $d \in \mathbb{R}^3$ from the cloud $\mathcal{C} \subset \mathbb{R}^3$, defined as:
\begin{equation}
\label{89:eq:3_1}
d \left( p, \mathcal{C} \right) = \min_{p' \in \mathcal{C}} 
\left\|
p' - p
\right\| \ ,
\end{equation}
where $\left\| p' - p \right\|$ is an Euclidean distance between points $p$ i $p'$.
Distance between point cloud $\mathcal{C}_1$ and $\mathcal{C}_2$ with a threshold $k$ is defined as:
\begin{equation}
\label{89:eq:3_10}
d_{k} \left( \mathcal{C}_1, \mathcal{C}_2 \right) 
=
\frac{1}{
\left| 
\mathcal{C}_1 \setminus \mathcal{O}_k
\right|}
\sum_{p \in \mathcal{C}_1 \setminus \mathcal{O}_k} 
d 
\left( 
p, \mathcal{C}_2
\right) \ ,
\end{equation}
where $\mathcal{O}_k$ is a set of points in a cloud $\mathcal{C}_1$ not having close neighbours in a cloud $\mathcal{C}_2$,
defined as:
\begin{equation}
\label{89:eq:3_11}
\mathcal{O}_k = 
\left\{
p \in \mathcal{C}_1 \ \ | \ \ d 
\left( 
p, \mathcal{C}_2
\right)
> k m
\right\}
\ ,
\end{equation}
where $m$ is a median distance between each point from the first cloud and the closest point from the second cloud. 
In the implementation of the presented method threshold $k=4$ was chosen.

Two face models being compared may have a different scale and orientation.
Scale difference is caused by the fact, that extrinsic parameters can be estimated for a sequence of images only up to an unknown scale factor. Thus a metric reconstruction is also possible up to a scale factor.
Orientation may be different because reconstructed head pose is aligned with the head pose on the first image.
In order to calculate a distance between two point clouds, they must be aligned first.
We use a variant of the popular ICP \footnote{\emph{ang.} Iterative Closest Point} \cite{89:rusin} algorithm, which can find a rigid body transformation aligning two point clouds. 

Let $C_s$ denotes a source point cloud and $C_d$ a destination point cloud.
Our modified version of ICP method has the following steps:

\begin{enumerate}
	
	\item Compute centroids of a source and destination cloud
	\begin{enumerate}
	  \item $\bar{c}_s = \left( \sum_{p \in \mathcal{C}_s} p \right) / |\mathcal{C}_s|$
	  \item $\bar{c}_d = \left( \sum_{q \in \mathcal{C}_d} q \right) / |\mathcal{C}_d|$
	\end{enumerate}
	
	\item Scale a source point cloud to match a destination cloud scale using a formula from \cite{89:horn}:
	
	\begin{enumerate}
		\item Compute scaling factor:
		$scale = \sqrt 
		\frac
		{\sum_{q \in \mathcal{C}_d} \left\| q_i - \bar{c}_d \right\|^2}
		{\sum_{p \in \mathcal{C}_s} \left\| p_i - \bar{c}_s \right\|^2}
		$
		\item Multiply coordinates of points in $\mathcal{C}_s$ by $scale$
	\end{enumerate}
	
	\item Align centroid of a source point cloud with a centroid of a destination cloud
	\begin{enumerate}
		\item Translate all point in $C_s$ by a vector $\bar{c}_d$ - $\bar{c}_s$.
	\end{enumerate}
		
	\item Choose a random sample $\mathcal{S} = \left\{ p_i \right\}$ of $s$ points from a source cloud $C_s$
	
	\item Match each point from a sample $S$ with the closest point in a destination cloud $C_d$. 
	Let $\mathcal{M} = \left\{ \left( p_i , q_i \right) \right\}$ denotes a set of corresponding point.
	\item Remove outliers from $\mathcal{M}$, that is remove pairs $(p_i,q_i)$  for which 
	$|d_i - q_i| > k m$, where $m$ is a median distance between pairs of corresponding points in $\mathcal{M}$, and $k$ is a small integer \footnote{In implementation $k=4$ was chosen.}.
	\item Find a rigid body transformation (rotation matrix $\mathbf{R}$ and translation vector $\mathbf{T}$) minimizing error metric $E \left( \mathbf{R}, \mathbf{T} \right)$ and apply the transformation on a source point cloud $\mathcal{C}_s$
		\item If number of iterations $< N$, go to point 4 else terminate the algorithm
\end{enumerate}

Algorithm parametrization and error metric $E$ were chosen experimentally to achieve good convergence and a reasonable running time.
Sample size $M$ is set to 500 (out of app. 40'000 points in clouds) and number of iterations $N=15$, as it was verified that larger values increase running time but do not improve convergence. 
As an error metric $E$, a point-to-plane error metric is chosen as it gives much faster convergence 
than a classic point-to-point error metric.
Point-to-plane error metric is given by the formula \cite{89:rusin}:
\begin{equation}
\label{89:eq:3:20}
E_{\mathrm{point-to-plane}} \left( \mathbf{R}, \mathbf{T} \right) = \sum_{i} 
\left(
\left(
\mathbf{R} p_i + \mathbf{T} - q_i
\right)
\cdot
n_i 
\right)
^ 2
\ ,
\end{equation}
where $n_i$ is a normal to the destination cloud surface at point $q_i$.

\section{Experiments}

\begin{figure}
\centering
\subfigure{
\includegraphics[height=2.6cm]{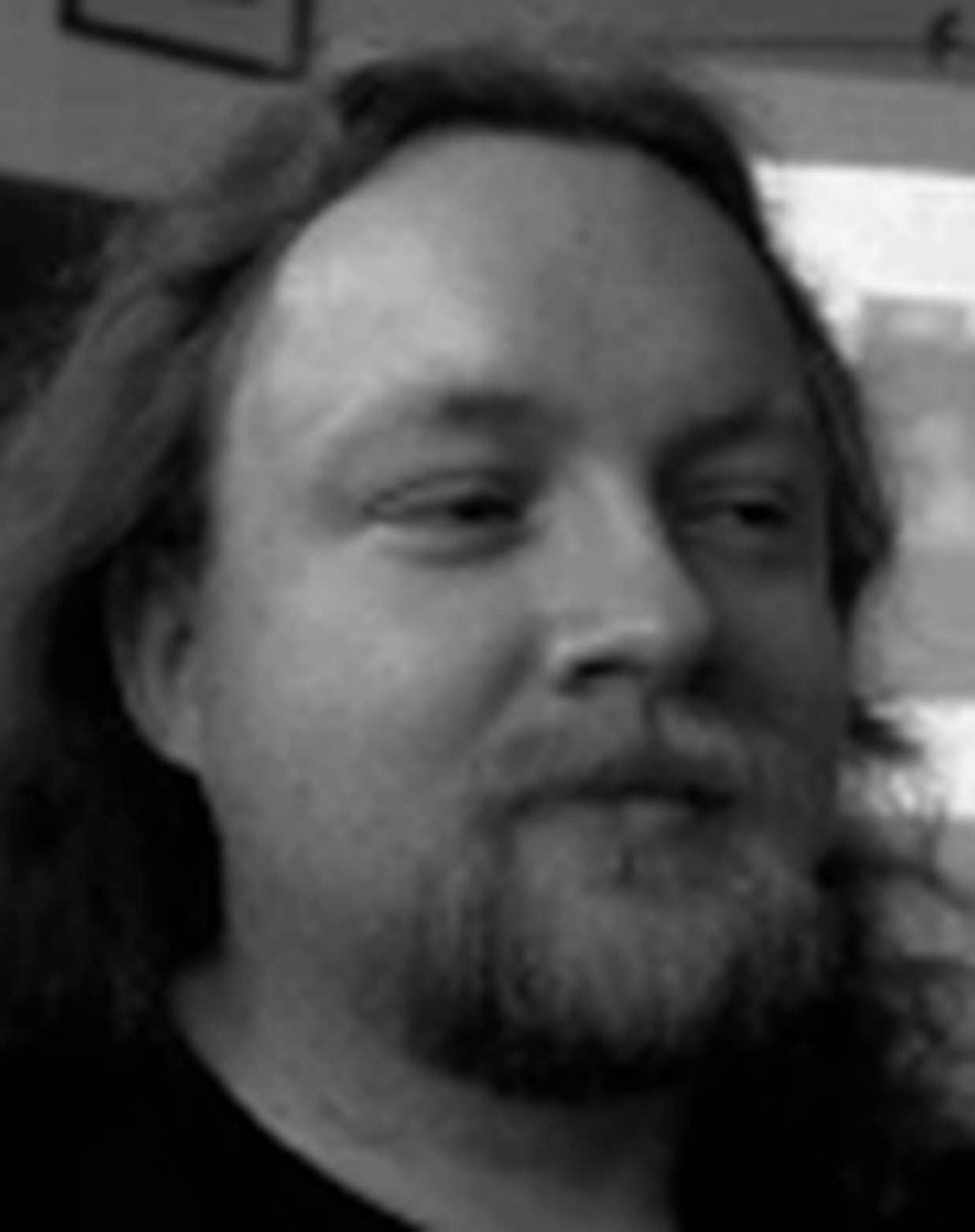}
}
\subfigure{
\includegraphics[height=2.6cm]{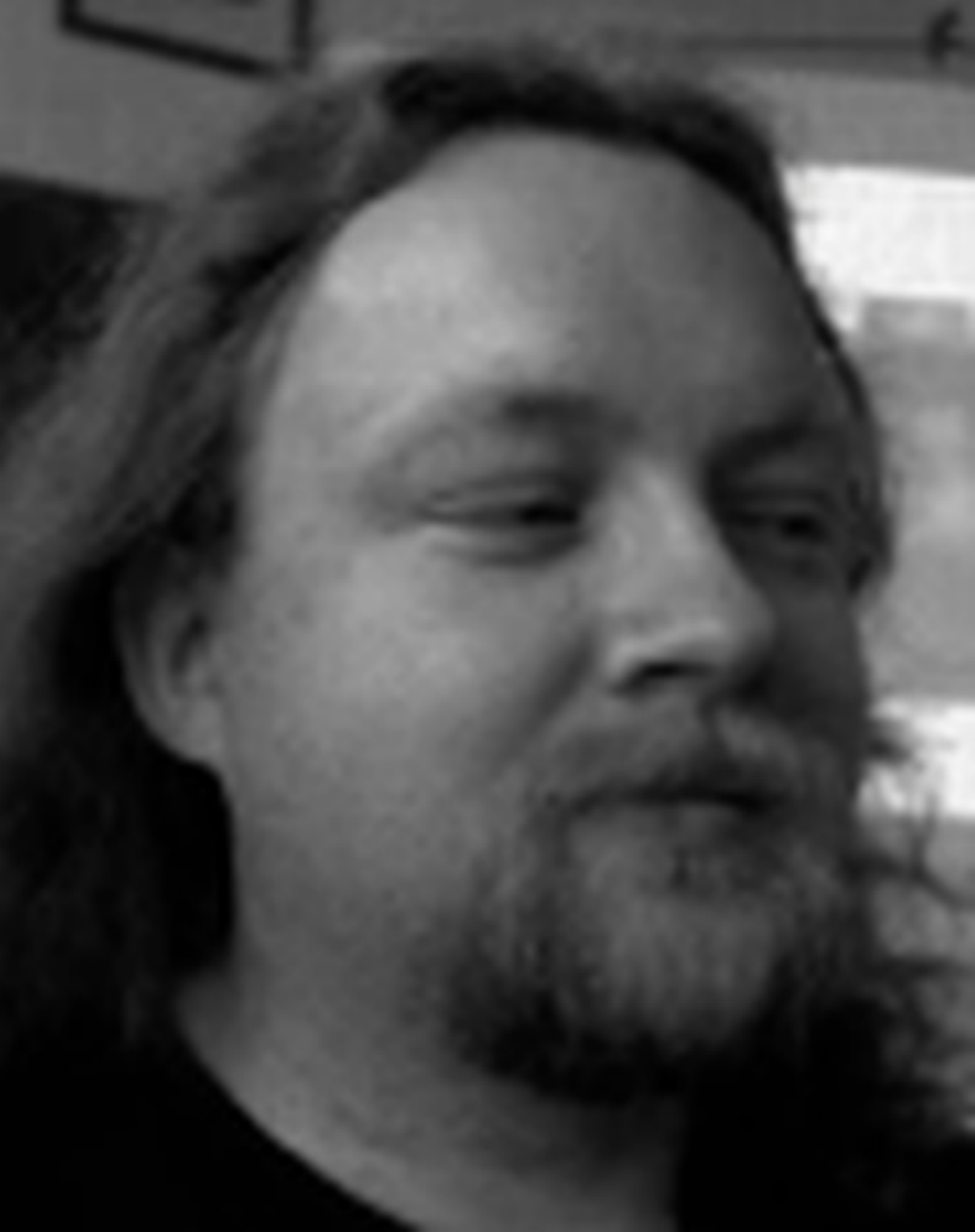}
}
\subfigure{
\includegraphics[height=2.6cm]{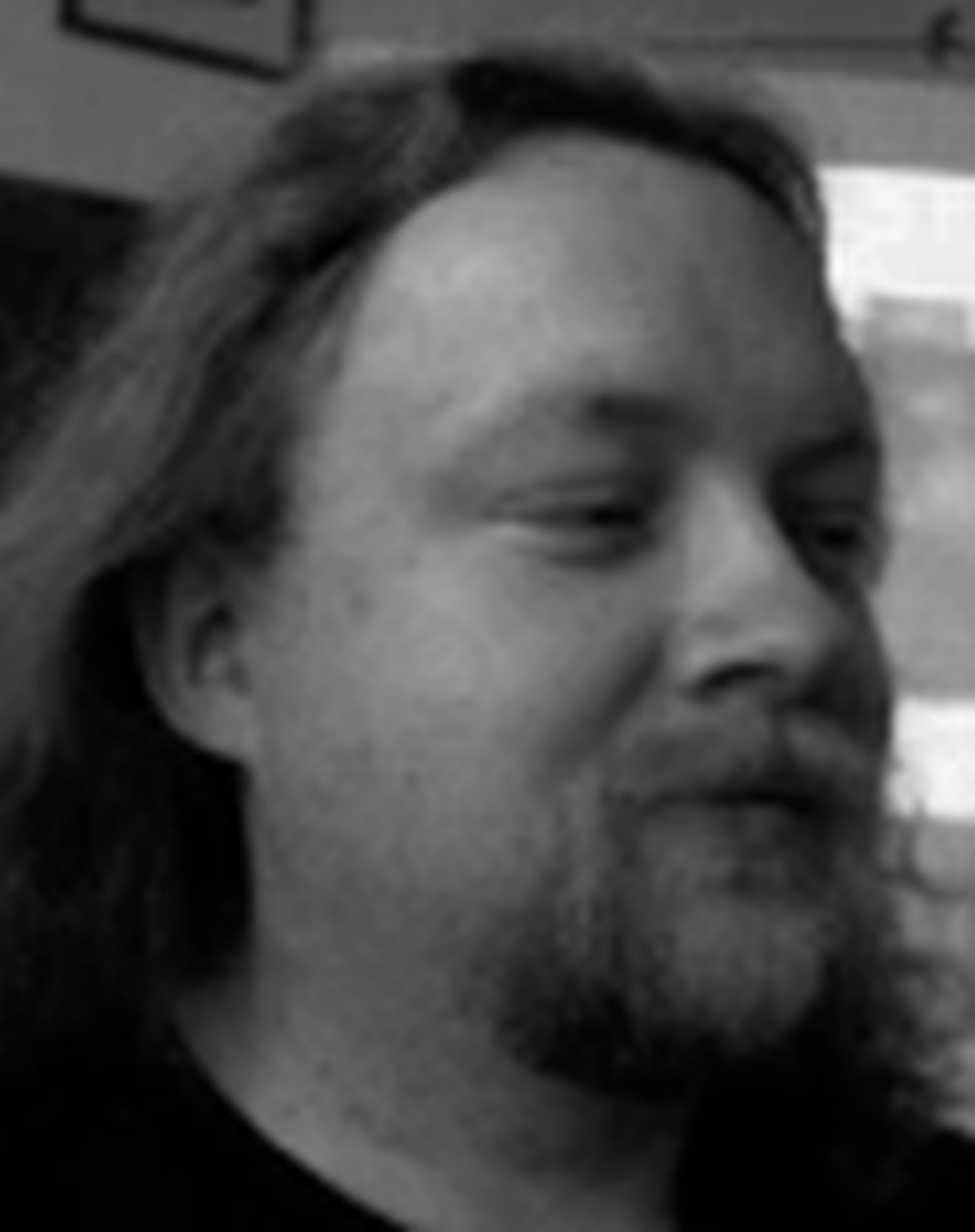}
}
\subfigure{
\includegraphics[height=2.6cm]{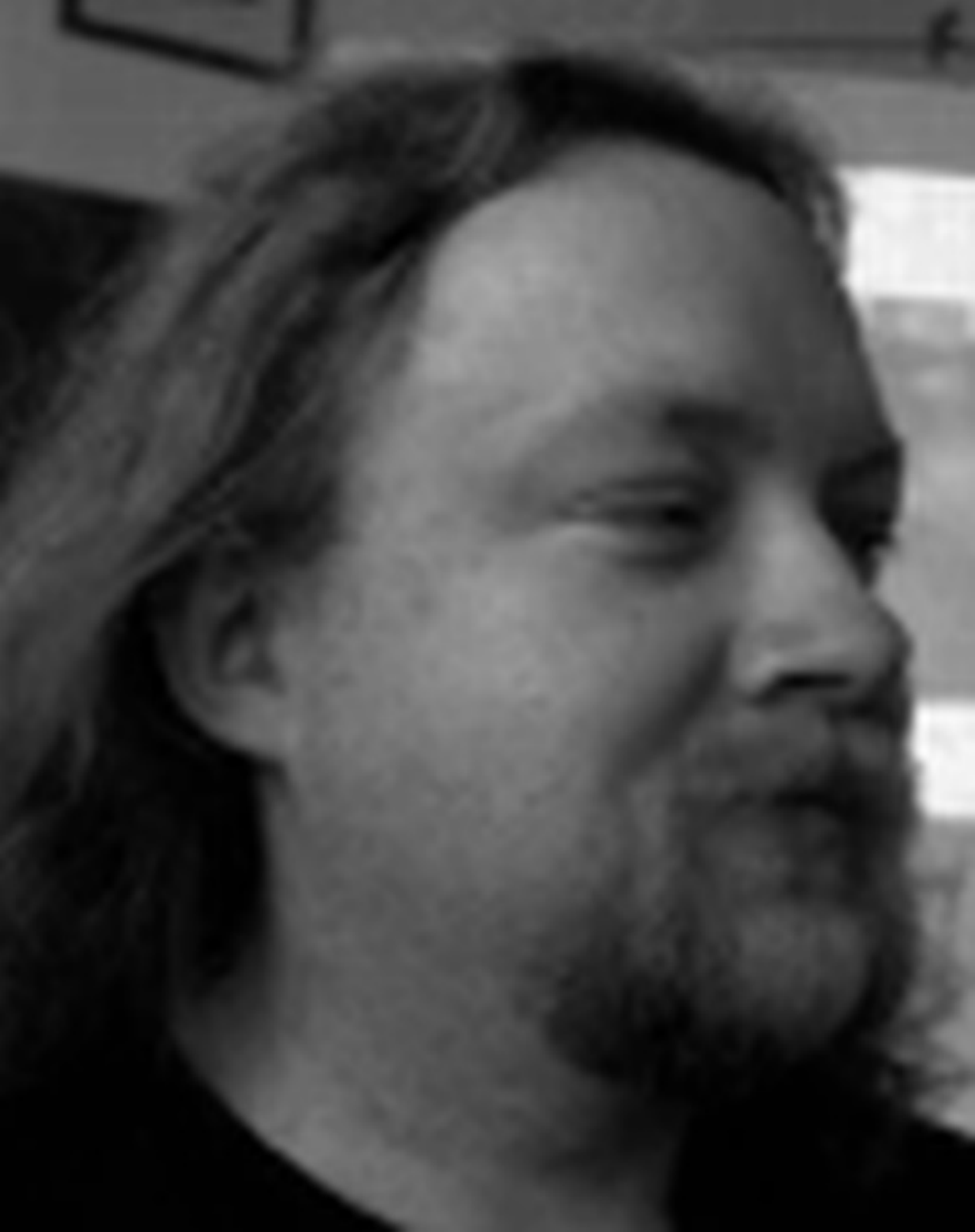}
}
\subfigure{
\includegraphics[height=2.6cm]{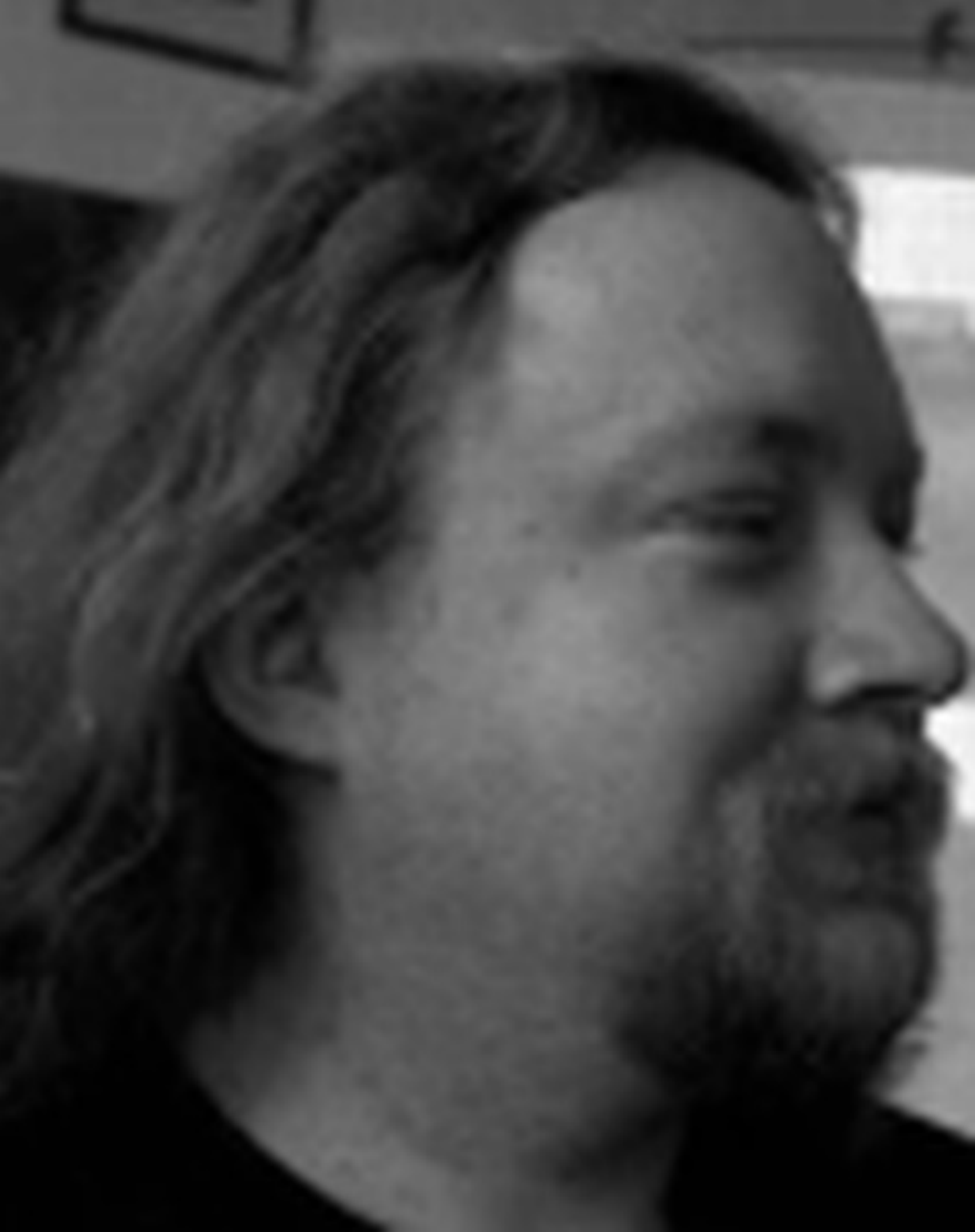}
}
\\
\subfigure{
\includegraphics[height=2.3cm]{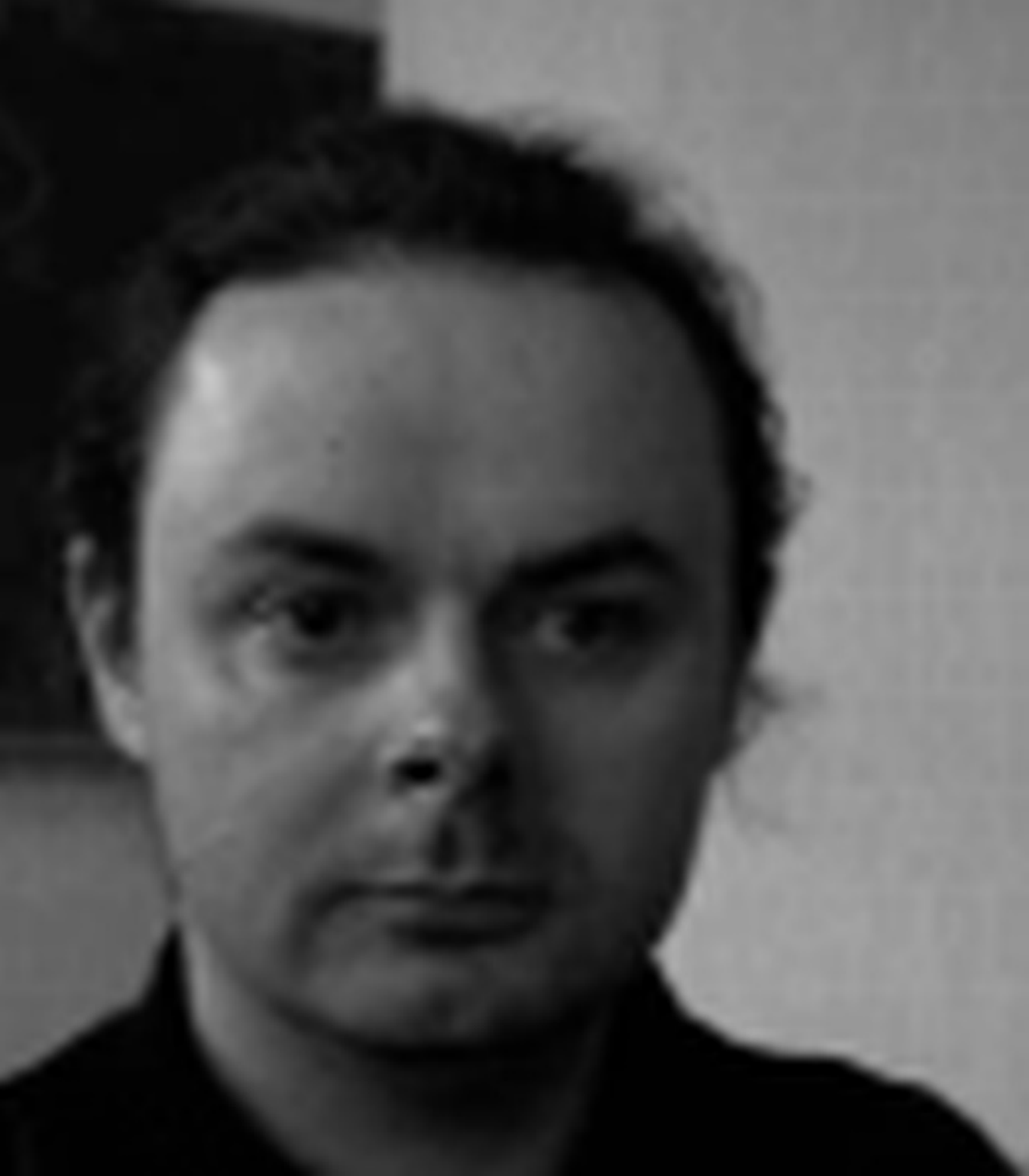}
}
\subfigure{
\includegraphics[height=2.3cm]{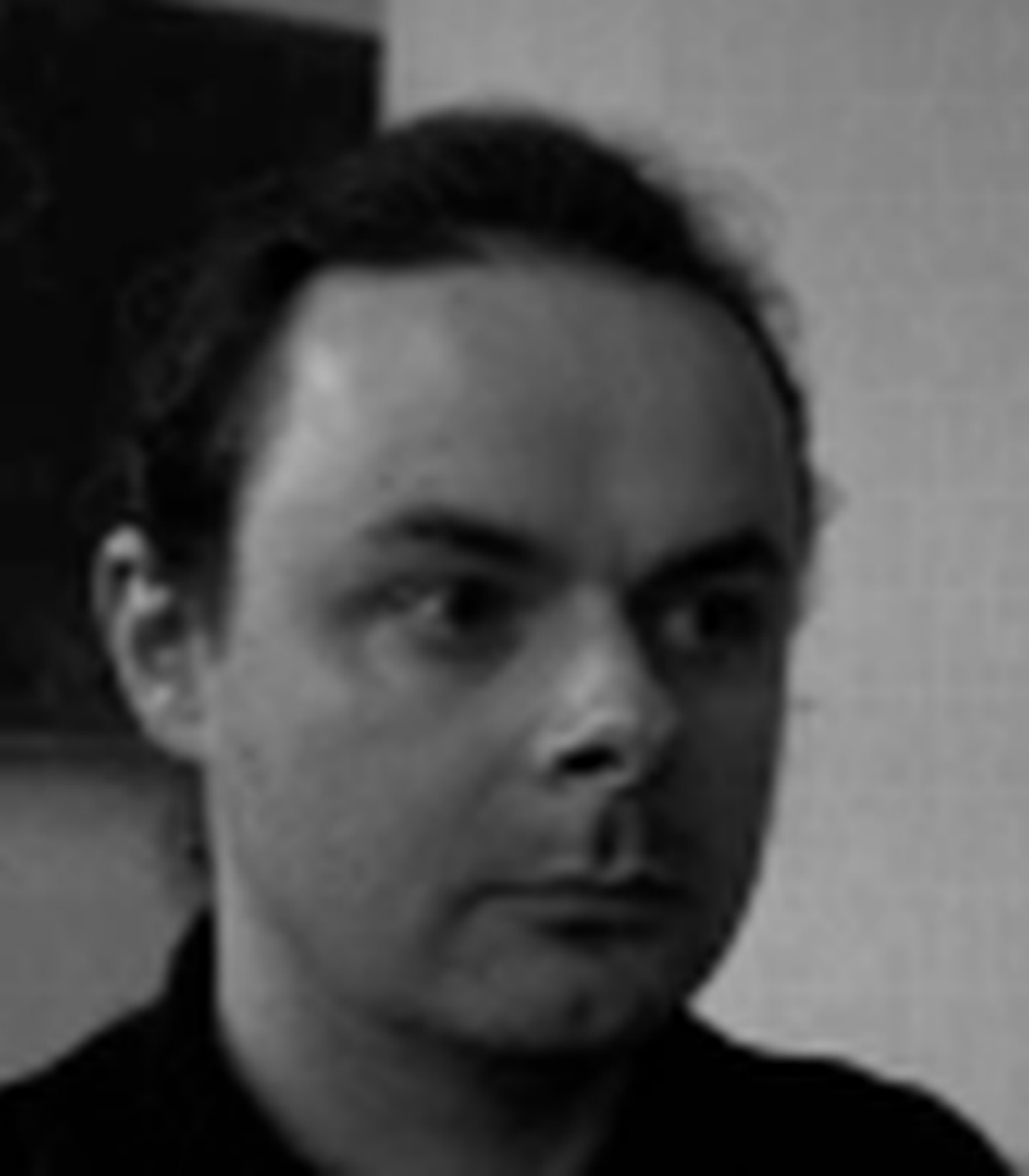}
}
\subfigure{
\includegraphics[height=2.3cm]{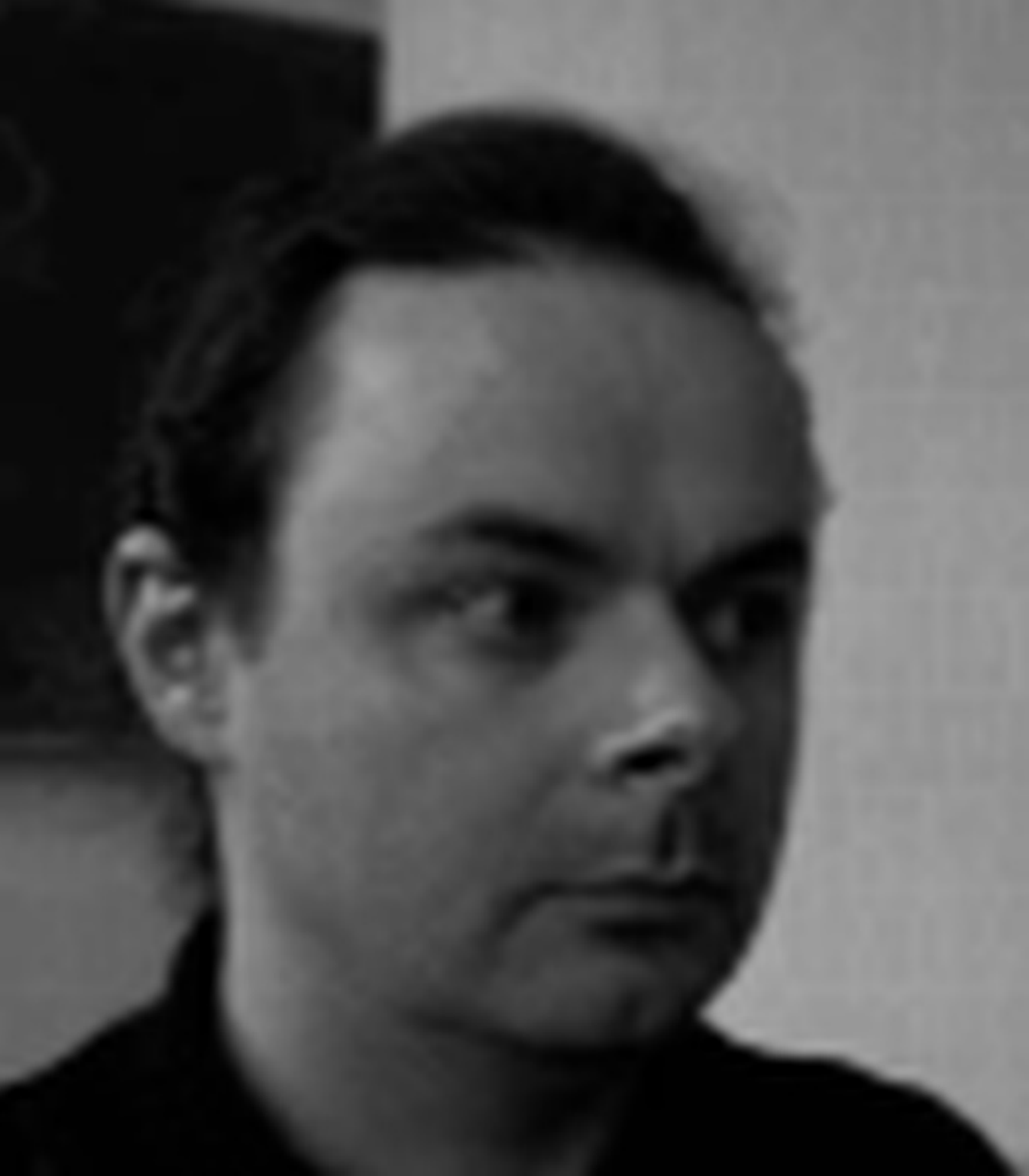}
}
\subfigure{
\includegraphics[height=2.3cm]{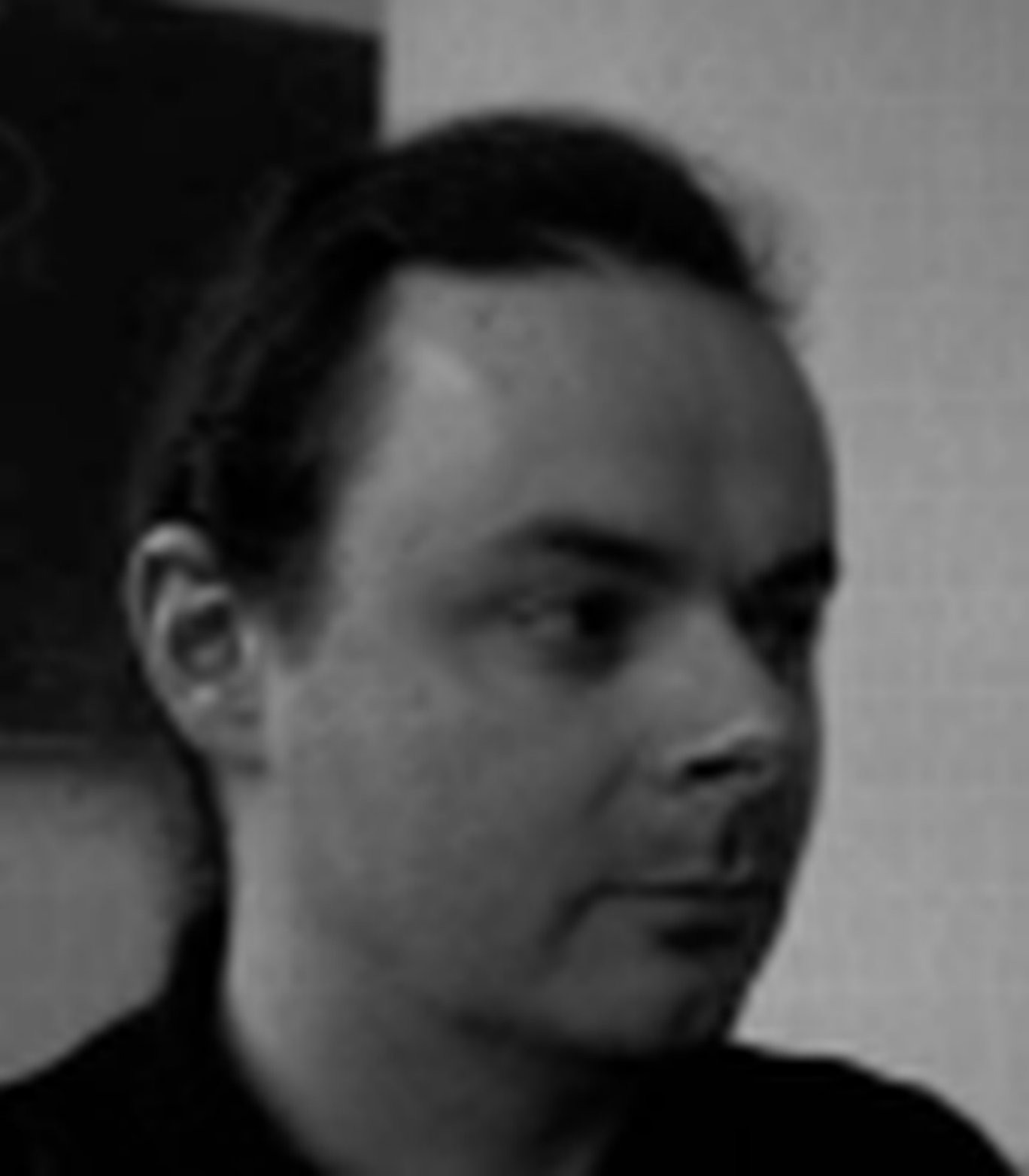}
}
\subfigure{
\includegraphics[height=2.3cm]{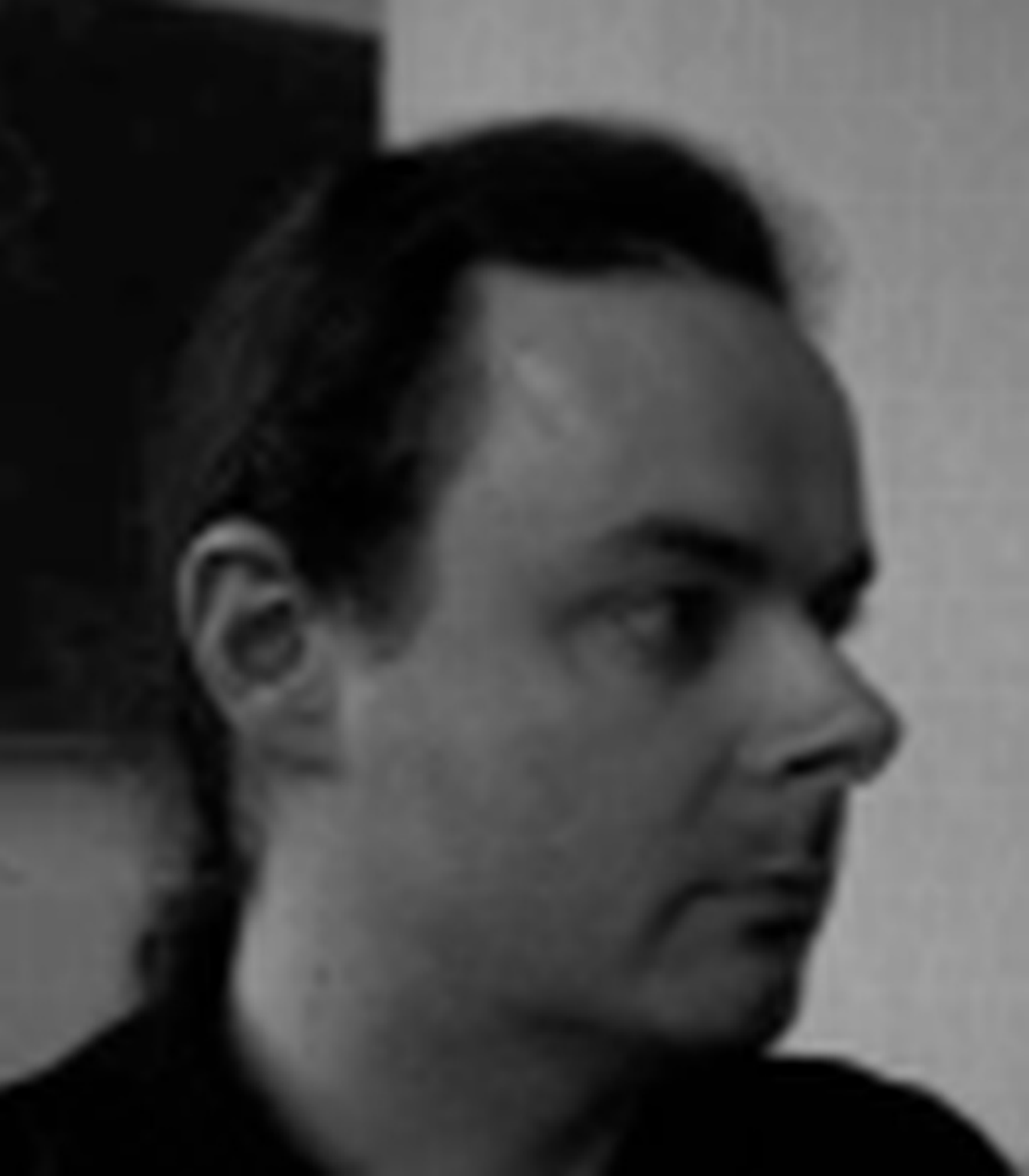}
}
\\
\subfigure{
\includegraphics[height=2.1 cm]{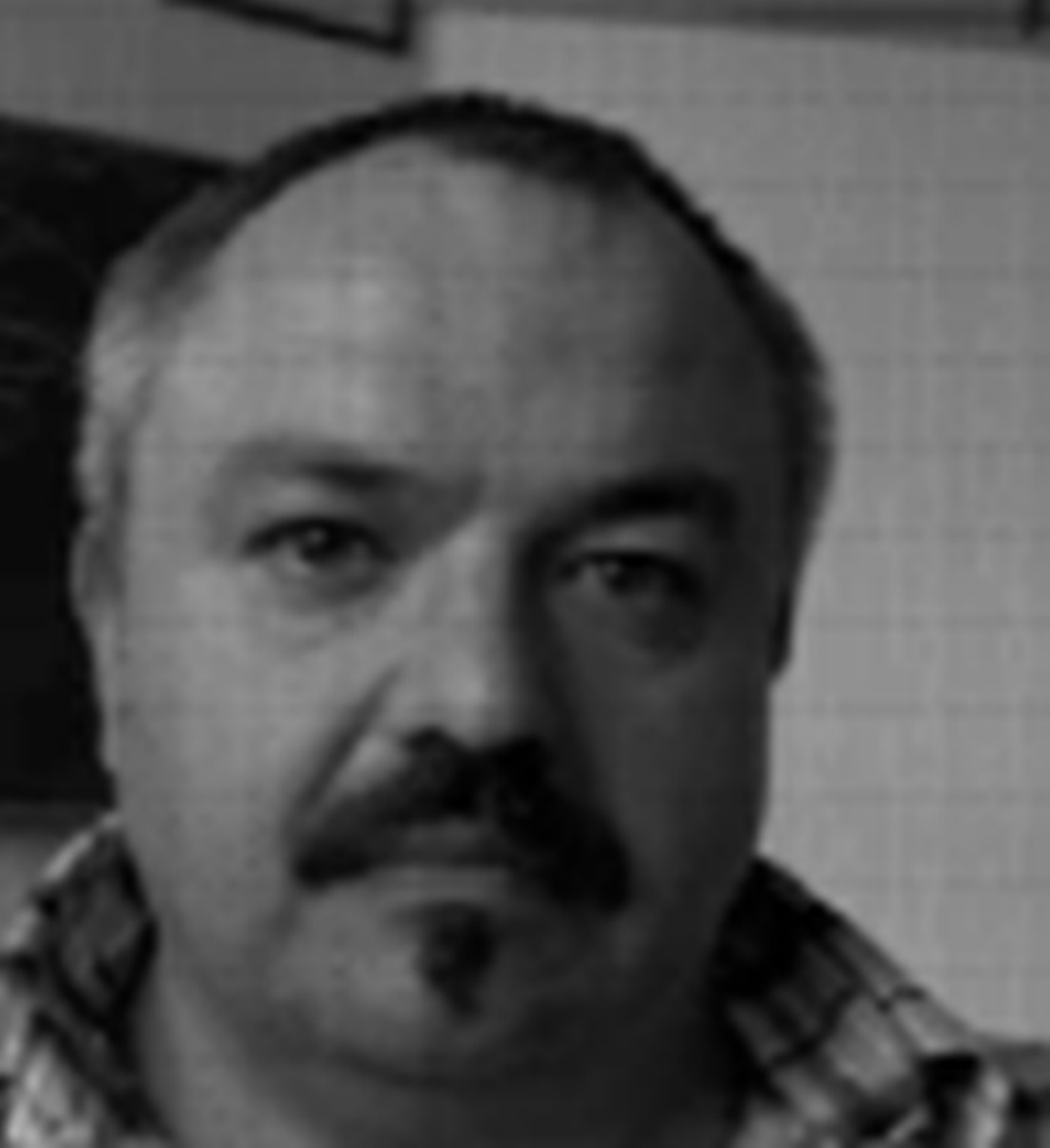}
}
\subfigure{
\includegraphics[height=2.1 cm]{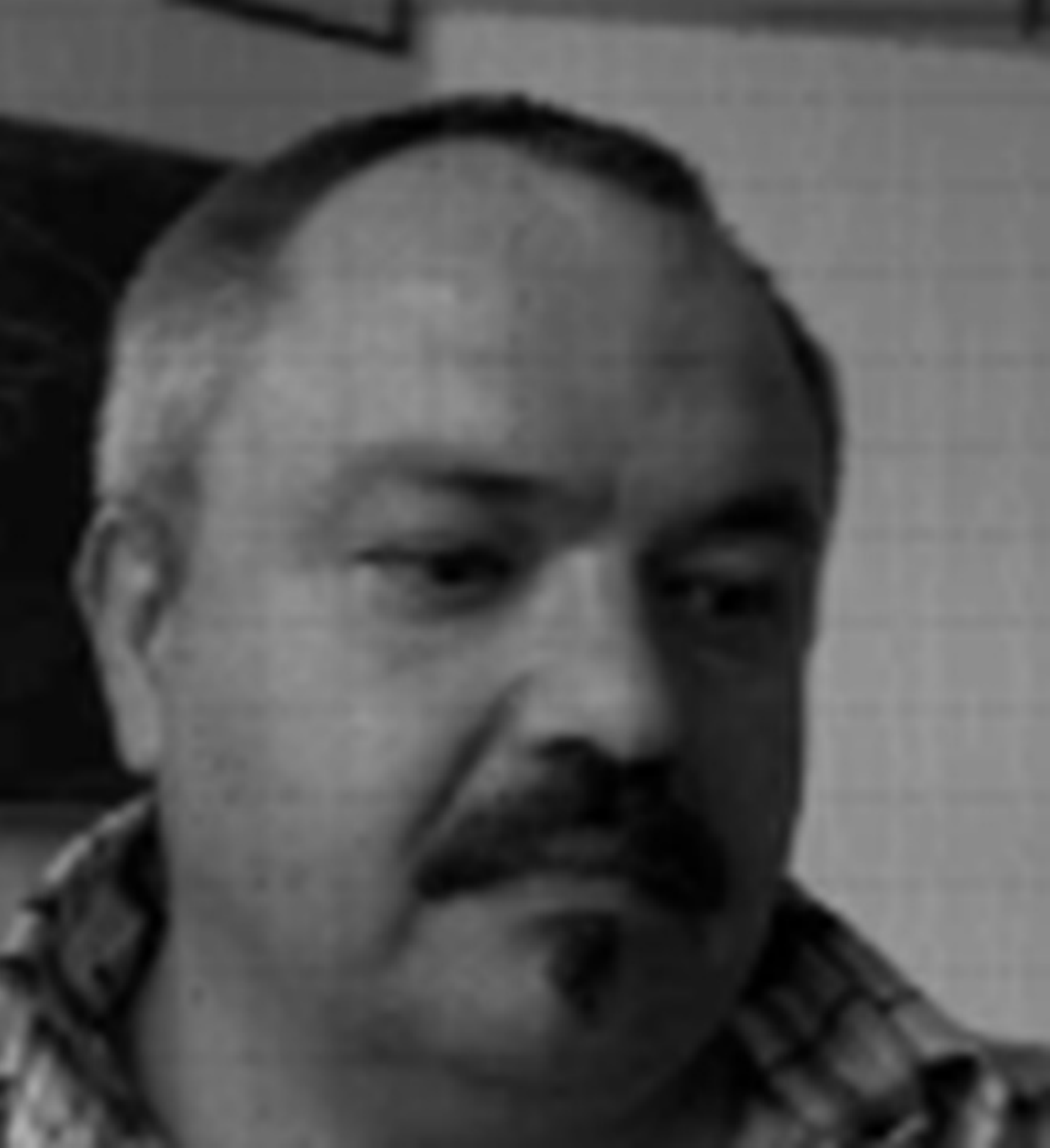}
}
\subfigure{
\includegraphics[height=2.1 cm]{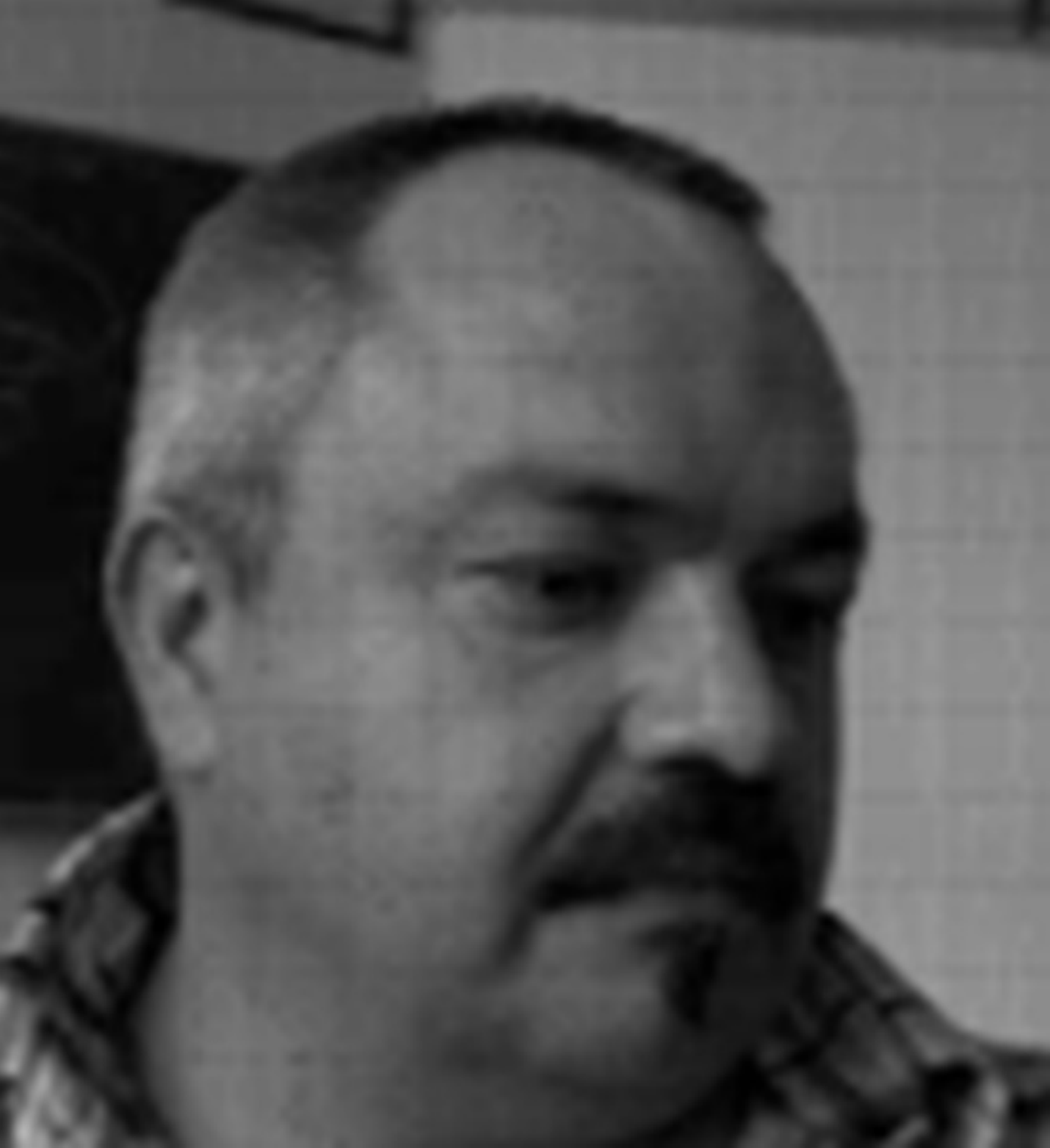}
}
\subfigure{
\includegraphics[height=2.1 cm]{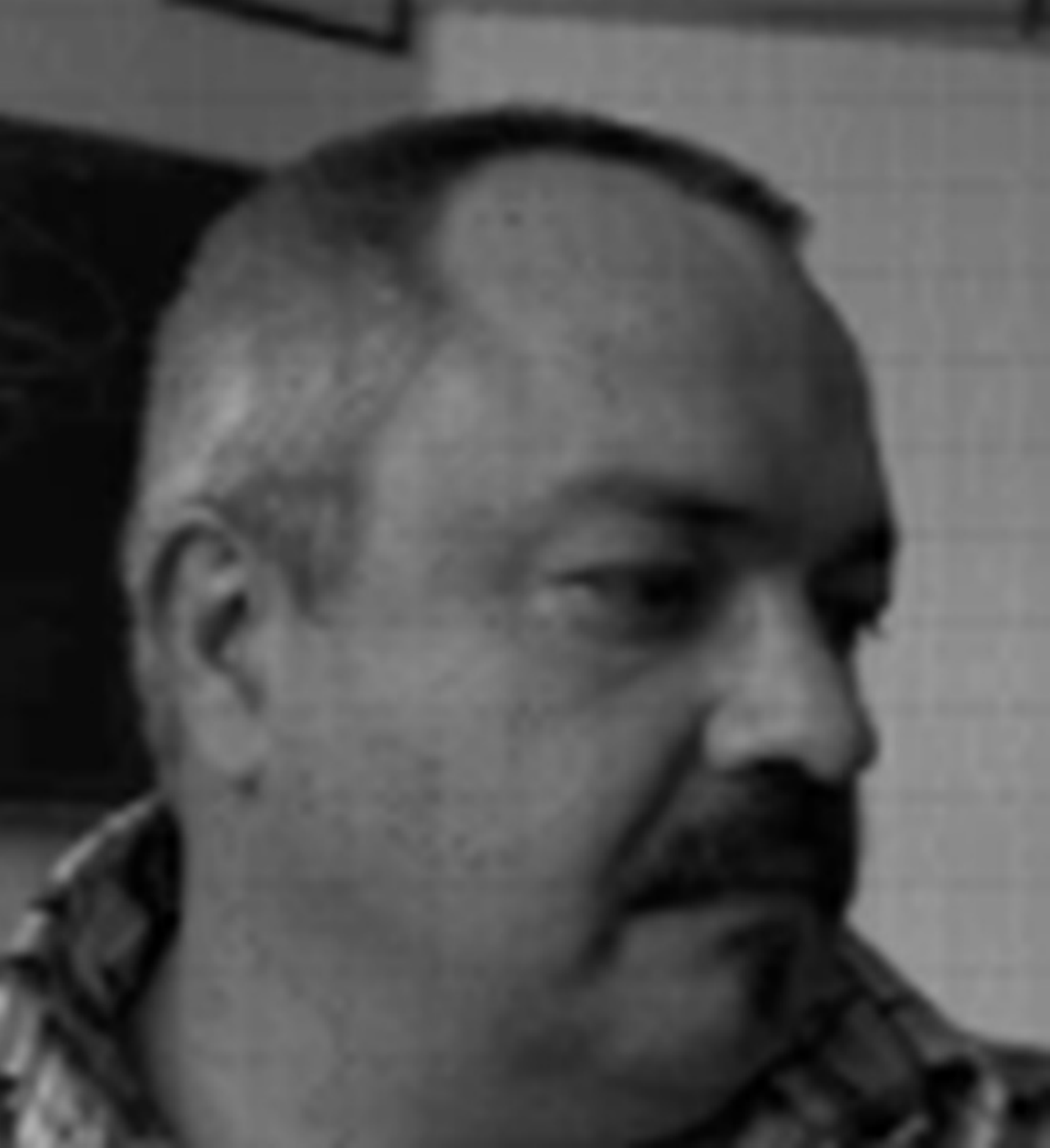}
}
\subfigure{
\includegraphics[height=2.1 cm]{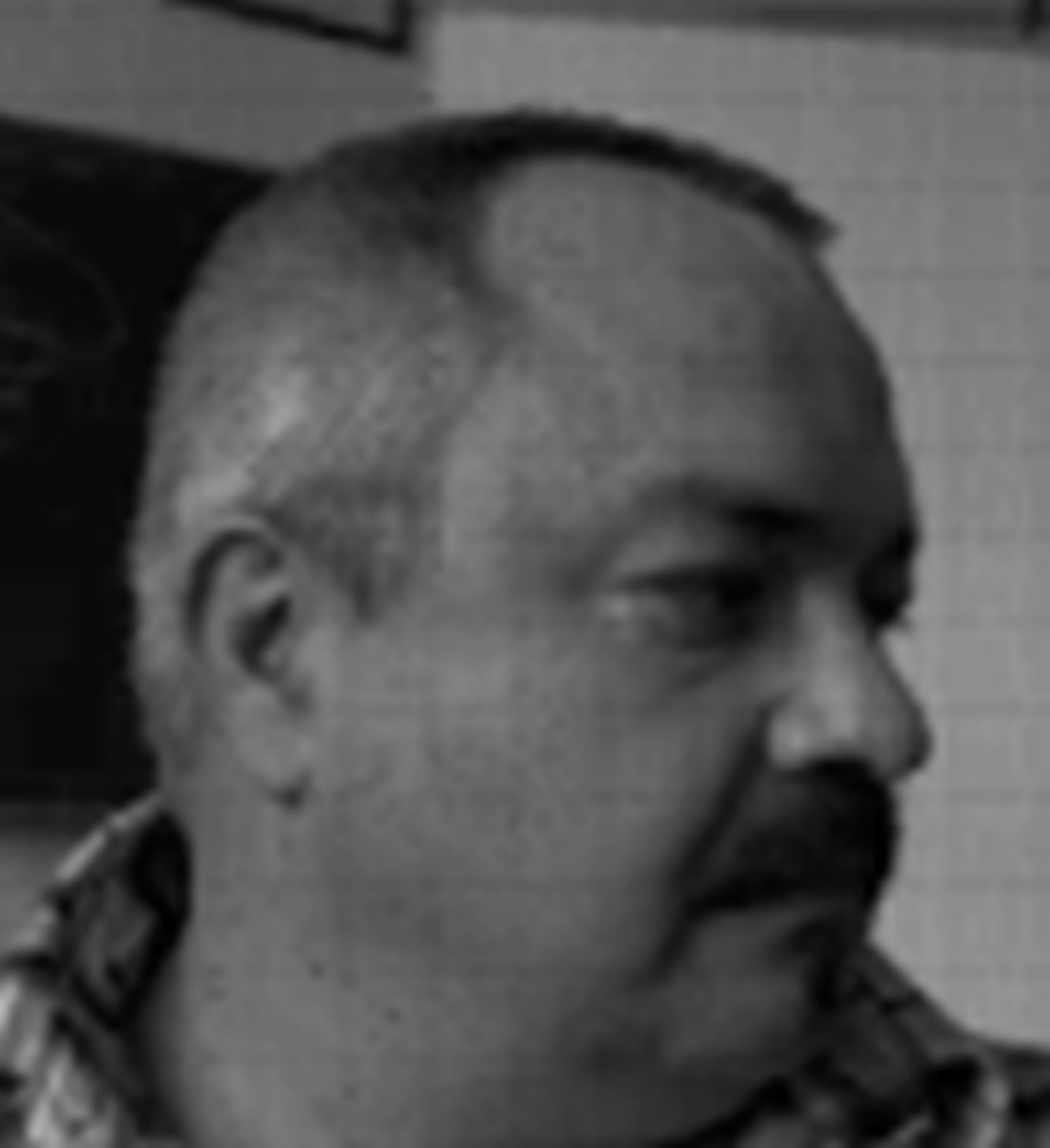}
}
\caption{Exemplery sequences from a test database.} 
\label{89:test_seq}
\end{figure}

This section presents results of an experimental verification of accuracy of the face recognition method presented in this paper.
Test database built by authors contains 81 image sequences of 27 individuals, 3 sequences per one person.
Images were acquired with Point Grey Chameleon camera 
\footnote{\url{http://www.ptgrey.com/products/chameleon/chameleon\_usb\_camera.asp}}
with 800x600 pixels resolution.
 In each sequence a persons sitting in front of a camera is asked to rotate his head right and left. Exemplary sequences are depicted on Fig. \ref{89:test_seq}. 
The database was split into 2 parts: 27 image sequences (1 per each individual) were used to build a gallery,
54 sequences (2 per each individual) were used to build a test set.

\paragraph{Error metrics} Face recognition system can be used to perform 2 tasks: verification and identification.
Verification is a task where the biometric system attempts to confirm an individual's claimed identity.
2 error metrics are used to assess accuracy of an identity verification task: FAR \footnote{False Acceptance Ratio} and FRR \footnote{False Rejection Ratio}.
FAR is defined as a ratio of a number of attempts when an identity was falsely positively verified to a number of all attempts.
FRR is defined as a ratio of a number of attempts when an identity was falsely negatively verified to a number of all attempts.

Identification is a task where biometric system searches a gallery for a reference matching submitted biometric sample, and if found, returns a corresponding identity.
Accuracy of identification tasks is measured with a CMC 
\footnote{Cummulative Match Characteristics}
curve. CMC is a function of a recognition rate as a number of best $n$-maches considered.
For a given $n$, recognition rate is a ratio of attempts when a chosen individual from a test set was among $n$ closest matches in the gallery to number of all attempts.
Clearly, when $n$ equals to the number of individuals in the gallery, recognition rate is equal to one.

\paragraph{Experiment 1} In this experiment accuracy of identity verification scenario was tested.
Each sequence from a test set was used to reconstruct a 3D face model which was matched against each face model in the gallery. If the distance between face model from a test set  
and a face model from a gallery was below a threshold $\Theta$ the identity was positively verified.
Otherwise identity was negatively verified.

\begin{figure}
	\centering
	\includegraphics[height=3.9 cm]{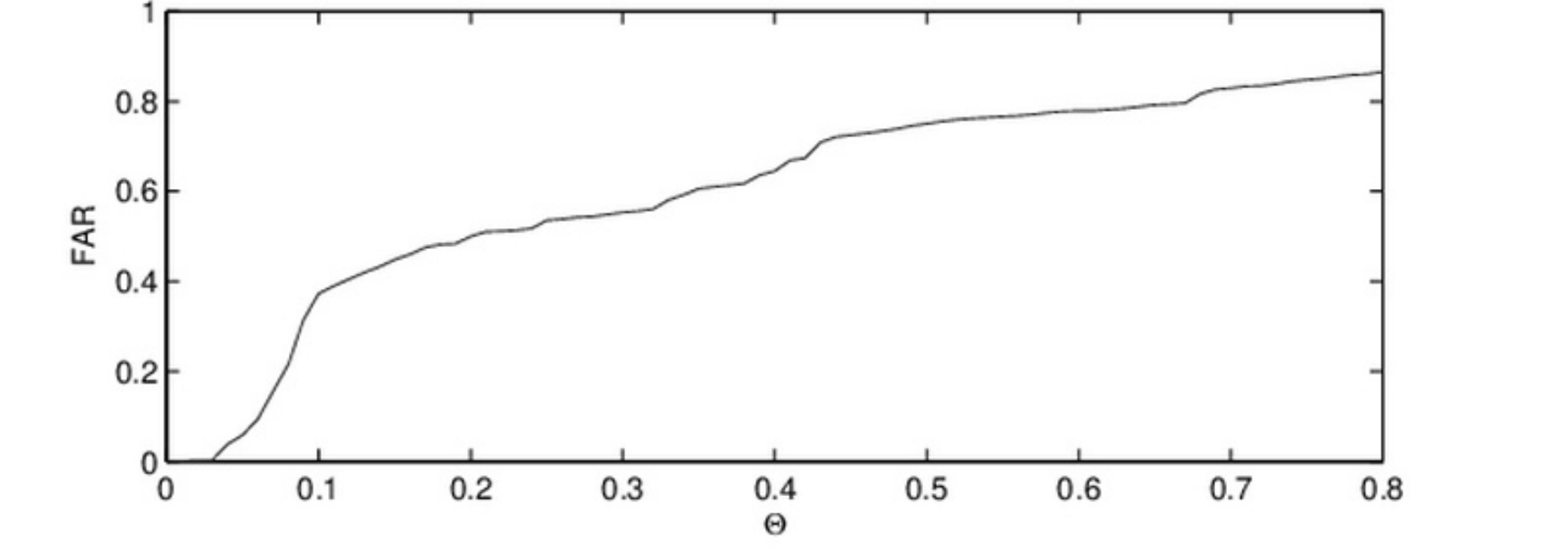}
	\caption{FAR as a function of a threshold $\Theta$.}
	\label{89:3_res_far}
\end{figure}

\begin{figure}
	\centering
	\includegraphics[height=3.9 cm]{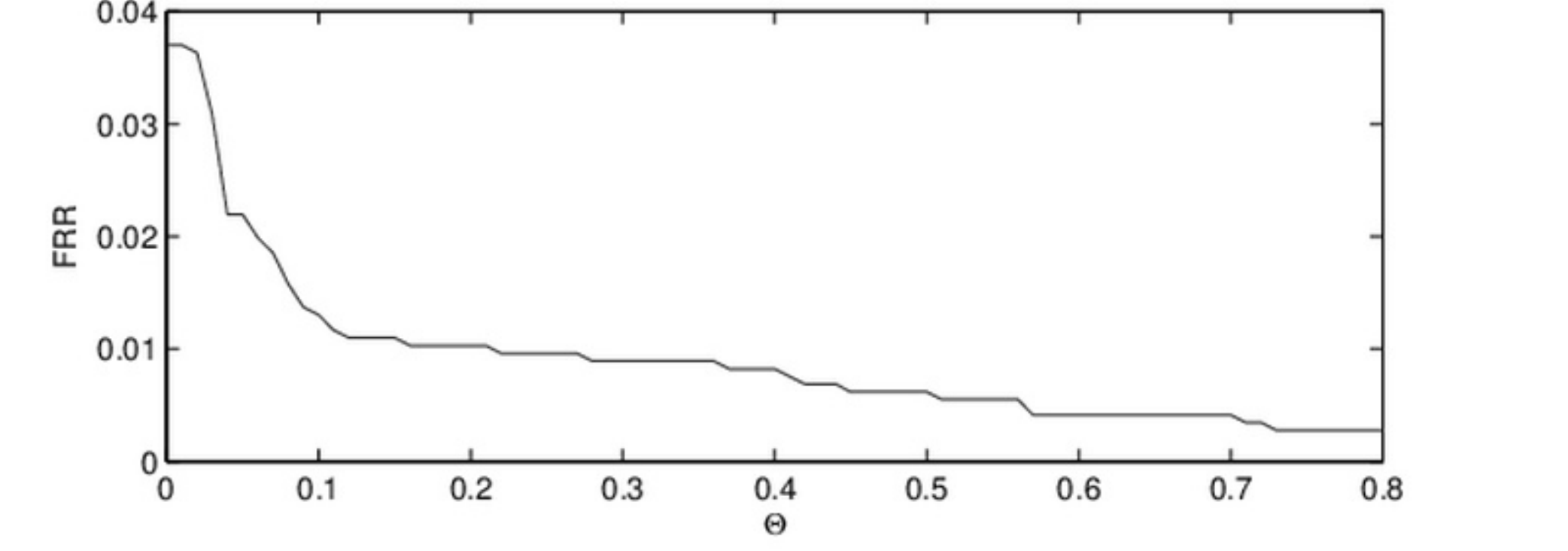}
	\caption{FRR as a function of a threshold $\Theta$.}
	\label{89:3_res_frr}
\end{figure}

Both FAR and FRR are dependent on threshold $\Theta$. When it's increased, more distant faces are identified as belonging to the same individual thus leading to FAR increase and FRR decrease. 
Fig. \ref{89:3_res_far} depicts values of FAR as a function of a threshold $\Theta$.
Fig. \ref{89:3_res_frr} shows values of FRR as a function of a threshold $\Theta$.

\begin{figure}
	\centering
	\includegraphics[height=3.9 cm]{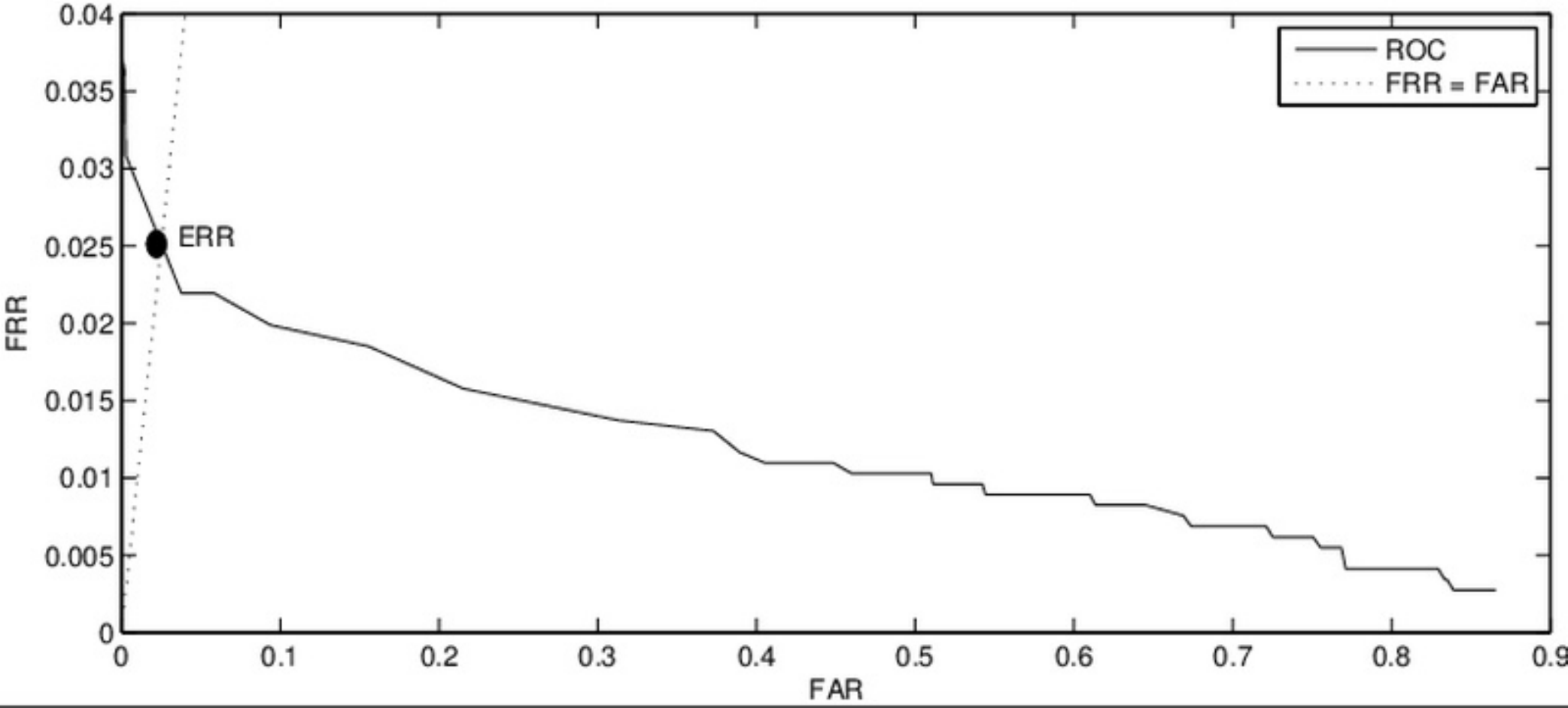}
	\caption{ROC curve and ERR point.}
	\label{89:3_res_roc}
\end{figure}

The trade-off between FAR and FRR rates is expressed using ROC
\footnote{Receiver Operating Characteristic}
curve and is  shown on Fig. \ref{89:3_res_roc}.
ERR
\footnote{Equal Error Rate},
that is a rate at which FAR = FRR is equal to $0.025$ and is a rather low value.
It means that in 2.5\% of attempts identity was falsely positively verified and in
2.5\% of attempts identify was falsely negatively verified.

\paragraph{Experiment 2} In this experiment identification in a  closed-set scenario was tested, as each individual from a test set was present in the gallery.
Each sequence from a test set was used to reconstruct a 3D face model which was matched against each face model in the gallery. 
Models with the closest distance were declared as a match.

\begin{figure}
	\centering
	\includegraphics[height=4.0 cm]{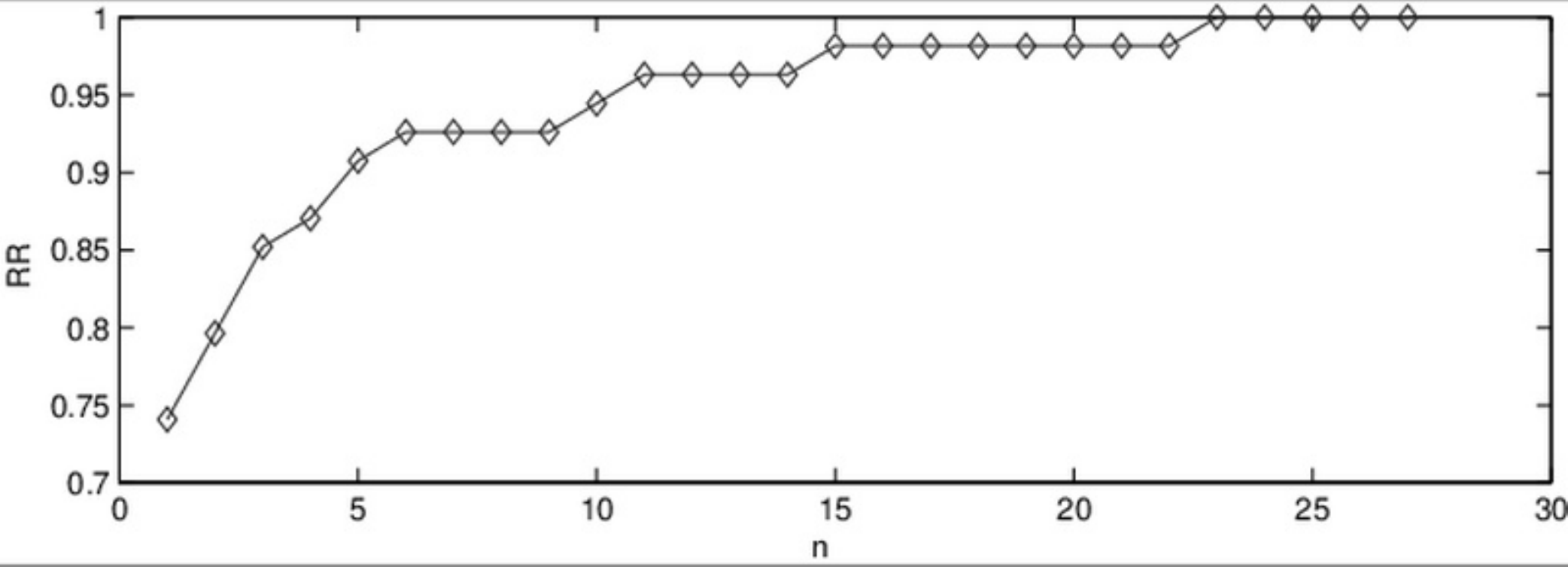}
	\caption{CMC curve}
	\label{89:3_res_cmc}
\end{figure}

Fig. \ref{89:3_res_cmc} shows resultant CMC curve.
When finding a single, best match in the gallery ($n=1$) for each individual from a test set, the method achieved almost 75\% accuracy.
If considering 5 best matches in the gallery ($n=5$), over 90\% accuracy was achieved.

\section{Conclusions and future work}

The presented method allows to achieve a reasonably good face recognition accuracy.
Although the results should be taken with care, as they were obtained using a relative small test database. 
To ensure validity of the proposed approach, experiments using much larger test database should be done.
For face recognition a relatively simple approach is used, based on direct comparison of two point clouds.
It's worth to investigate more advanced approaches, e.g. based on comparison of a local characteristics such as nose profile, or a relative position of eyes, nose  and lips.
It must be noted that recognition is based only on spatial information and 2D information (texture) is not used.
Combining 2 modalities (shape and texture) may allow to achieve better recognition rates.

Face reconstruction method presented in this paper consists of two separate and distinct steps. 
Sparse point cloud build during the process of estimating extrinsic parameters is discarded, and only
extrinsic paratmers are passed to the second step (multi view stereo reconstruction).
Potentially 3D points from a sparse point cloud created in the first step can be used to initialise multi-view stereo reconstruction process.

\end{document}